\documentclass[11pt]{article}

\usepackage[final]{acl}

\usepackage{times}
\usepackage{latexsym}

\usepackage[T1]{fontenc}
\usepackage[utf8]{inputenc}

\usepackage{microtype}

\usepackage{inconsolata}

\usepackage{graphicx}
\usepackage{subcaption}
\usepackage{booktabs}
\usepackage{multirow}
\usepackage{siunitx}
\usepackage{amsmath,amssymb}
\usepackage{threeparttable}
\usepackage{placeins}
\usepackage[linesnumbered,ruled,vlined]{algorithm2e}
\usepackage{makecell}
\usepackage{enumitem}
\usepackage{pifont}

\newcommand{\sysname}{\texttt{GraphDx}}

\title{GraphDx: A Cost-Aware Knowledge-Enhanced Multi-Agent Framework for Sequential Diagnosis}

\author{
 \textbf{Shaoting Tan\textsuperscript{1}},
 \textbf{Ning Liu\textsuperscript{1}},
 \textbf{Yuntao Du\textsuperscript{1}},
 \textbf{Shuyue Wei\textsuperscript{1}},
\\
 \textbf{Wu Shuai\textsuperscript{2}},
 \textbf{Qian Li\textsuperscript{1}},
 \textbf{Yanyu Xu\textsuperscript{1}},
 \textbf{Wei Zhang\textsuperscript{3}},
\\
 \textbf{Lizhen Cui\textsuperscript{1}},
 \textbf{Haitao Yuan\textsuperscript{4}},
\\
\\
 \textsuperscript{1}Shandong University,
 \textsuperscript{2}Nankai University,
 \textsuperscript{3}East China Normal University,
 \textsuperscript{4}National Technological University,
\\
 \small{
   \textbf{Correspondence:} \href{mailto:email@domain}{liun21cs@sdu.edu.cn}
 }
}

\begin{document}
\maketitle

\begin{abstract}
  % Sequential diagnosis requires balancing accuracy and cost through multi-turn interactions. Existing Large Language Model (LLM) systems often suffer from a "Knowledge-Reasoning Mismatch," leading to excessive testing and poor interpretability due to the lack of explicit structural guidance.
  % To address this, we propose \sysname{}, a fully automated knowledge-enhanced framework. It first utilizes LLMs to construct a Medical Diagnosis Knowledge Graph (MDKG) with typicality associations and test utilities. Then, it drives a graph-augmented agent where the LLM handles natural language interaction, while a graph-based inference engine on the graph guides differential diagnosis and cost-sensitive test recommendation.
  % Experimental results on MedQA and MIMIC-IV datasets demonstrate that \sysname{} significantly outperforms standard LLMs across multiple base models (DeepSeek-V3, Kimi-k2, Llama-3.3). Specifically, it improves diagnostic success rates to nearly 90\% on MedQA and reduces average test costs by approximately 40\%, offering a robust, economical, and interpretable solution for automated medical diagnosis. 
  % To facilitate future research, we release our code, constructed MDKGs, and the simulation environment at \url{https://anonymous.4open.science/r/GraphDx-BB8B}.
Sequential diagnosis requires balancing diagnostic accuracy against resource costs through iterative information gathering. Existing Large Language Model (LLM) approaches exhibit a critical \emph{knowledge-reasoning gap}: despite encoding extensive medical knowledge, they struggle to reason systematically under cost constraints, often resorting to excessive testing.
  We propose \sysname{}, a knowledge-enhanced framework with two core innovations. First, we design an automated pipeline that leverages LLMs to construct Medical Diagnosis Knowledge Graphs (MDKGs) with quantized typicality, action-centric topology, and dual-objective attributes for both diagnostic relevance and cost-sensitivity. Second, we introduce three collaborative agents (Perception, Reasoning, and Decision) where the Perception and Decision Agents handle language understanding and generation, while the Reasoning Agent performs deterministic evidence scoring and cost-aware planning on the MDKG.
  Experiments on MedQA and MIMIC-IV across three LLM backbones (DeepSeek-V3, Kimi-k2, Llama-3.3) show that \sysname{} improves diagnostic success rates from 50--68\% to 79--93\% while reducing test costs by 20--54\%, providing a robust, economical, and interpretable solution for automated clinical diagnosis.

\end{abstract}

\section{Introduction}
\label{sec:intro}

Large Language Models (LLMs) have achieved remarkable success on medical examinations, with models like GPT-4 surpassing human performance on USMLE~\cite{nori2023capabilitiesgpt4medicalchallenge, singhalLargeLanguageModels2023}, demonstrating that LLMs encode extensive medical knowledge within their parameters. However, translating this success to real clinical practice, particularly \textbf{sequential diagnosis}, remains a significant challenge.

Unlike static question answering, sequential diagnosis mirrors actual clinical workflows: physicians iteratively gather information through symptom inquiries and diagnostic tests, making decisions under uncertainty while balancing diagnostic accuracy against resource costs including monetary expense, patient discomfort, and time. This multi-turn process requires not merely knowledge recall, but systematic reasoning that coordinates information acquisition with hypothesis refinement.

Current LLM-based approaches exhibit a critical limitation we term the \emph{knowledge-reasoning gap}: despite possessing substantial medical knowledge, LLMs struggle to reason systematically under cost constraints. Recent studies~\cite{nori2025sequentialdiagnosislanguagemodels, jia2025ddodualdecisionoptimizationllmbased} reveal that LLM agents tend toward ``defensive medicine,'' ordering excessive tests to compensate for reasoning uncertainty while failing to leverage discriminative features for efficient differential diagnosis. This behavior stems from the implicit, unstructured nature of LLM reasoning, which lacks explicit domain models enabling systematic hypothesis testing.
We identify two key challenges underlying this gap:

\paragraph{Challenge 1: Diagnosis-Ready Knowledge Graph Construction.}
Sequential diagnosis requires \emph{quantitative} attributes beyond semantic relations, including symptom typicality and test utilities. Existing knowledge graphs like UMLS lack these attributes, while current automated methods~\cite{chen2023autokgefficientautomatedknowledge, anuyah-etal-2025-automated} extract only semantic triples. Moreover, medical concepts exhibit context-dependent semantics (e.g., ``heart failure'' as diagnosis vs. complication), requiring dynamic role modeling that static ontologies cannot provide.

\paragraph{Challenge 2: Effective Multi-Agent Collaboration.}
Bridging neural language understanding with symbolic graph reasoning requires orchestrating observation extraction, graph grounding, evidence scoring, and response generation. Recent approaches like MAI-DxO~\cite{nori2025sequentialdiagnosislanguagemodels} rely on prompt-based coordination, but their complexity demands strong instruction-following from base models, causing performance degradation on less capable LLMs.

\begin{figure*}[htbp]
  \centering
  \includegraphics[width=\linewidth]{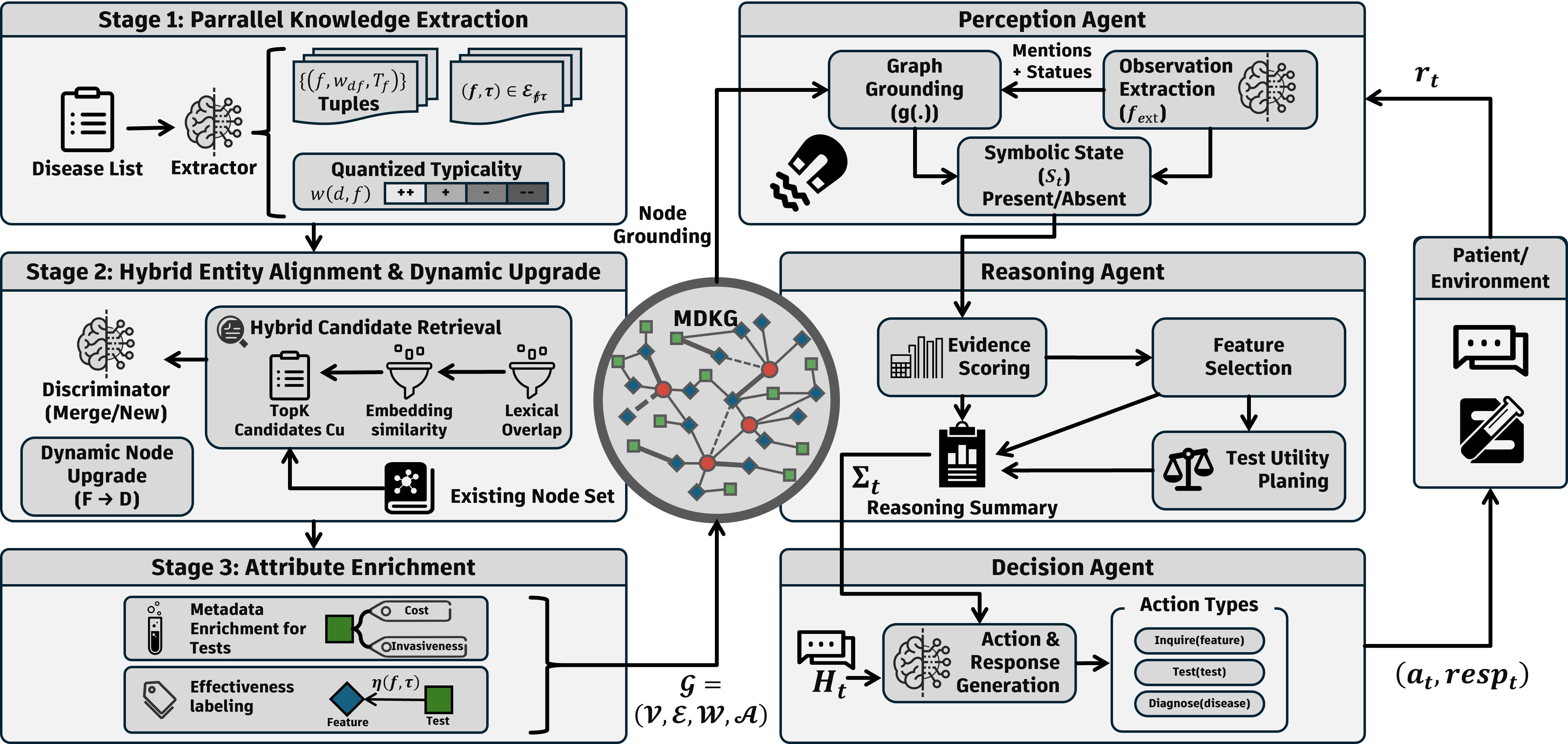}
  \caption{Overview of the \sysname{} framework. Left: Automated MDKG Construction Pipeline (Algorithm 1), utilizing parallel extraction and hybrid alignment to build the graph. Right: Graph-Augmented Diagnostic Agent (Algorithm 2), employing a knowledge-enhanced loop of observation extraction, graph grounding, evidence scoring, and utility-based action planning.}
  \label{fig:overview}
  \vspace{-5mm}
\end{figure*}

To address these challenges, we propose \sysname{}, a knowledge-enhanced framework comprising two integrated components (Figure~\ref{fig:overview}):

\noindent \textbf{(1) Automated MDKG Construction} (\S\ref{sec:mdkg_construction}) addresses \textbf{Challenge 1}. We design an automated pipeline where specialized agents collaborate to construct MDKGs enriched with \emph{quantitative diagnostic attributes} including typicality weights and test cost-effectiveness metrics that existing knowledge graphs lack. The pipeline employs a three-stage process: parallel knowledge extraction with quantized typicality, hybrid entity alignment with dynamic node upgrade for handling semantic role transitions, and attribute enrichment for cost-sensitive planning. This enables complex causal chains to emerge automatically without manual annotation.

\noindent \textbf{(2) Collaborative Diagnostic Agents} (\S\ref{sec:reasoning}) addresses \textbf{Challenge 2}. We introduce three specialized agents: a \emph{Perception Agent} that extracts observations from patient responses and grounds them to graph nodes, a \emph{Reasoning Agent} that performs deterministic evidence scoring and cost-sensitive planning on the MDKG, and a \emph{Decision Agent} that generates clinically appropriate responses. This architecture offloads complex reasoning to the graph, reducing dependence on base model capabilities.

We evaluate \sysname{} on MedQA-Extended and MIMIC-IV datasets across three LLM backbones. Results demonstrate that \sysname{} improves diagnostic success rates from 50\% to over 88\% while reducing test costs by 20\% to 50\%, with consistent gains across all tested models, including scenarios where sophisticated multi-agent baselines fail. To support reproducibility and further research, we open-source our implementation, including the automated graph construction pipeline and the \texttt{ClinicSim} environment at \url{https://anonymous.4open.science/r/GraphDx-BB8B}.

\section{Related Work}
\label{sec:related}
\newcommand{\cmark}{\ding{51}} % ✓
\newcommand{\xmark}{\ding{55}} % ✗
\newcommand{\wmark}{\ensuremath{\triangle}} % △
\newcommand{\naMark}{\textemdash} % —

\begin{table}[htbp]
  \centering
  \caption{Comparison of Recent Works and Ours.\\\cmark=Yes, \xmark=No, \wmark=Weak/partial, \naMark=Not applicable.}
  \label{tab:rw-compare}
  \resizebox{\linewidth}{!}{
    \begin{tabular}{l|c|c|c|c}
      \toprule
      Method                                & \makecell{Explicit                                                       \\ KG}  & \makecell{Auto                               \\ Construction} & \makecell{Sequential \\ Diagnosis} & \makecell{Cost- \\ Sensitive} \\
      \midrule
      \makecell{End-to-End LLM~                                                                                        \\ \cite{singhalLargeLanguageModels2023}}                       & \xmark                 & --           & \wmark         & \xmark           \\ \hline
      \makecell{Multi-Agent LLM~                                                                                       \\ \cite{nori2025sequentialdiagnosislanguagemodels}} & \xmark                 & --           & \cmark          & \wmark         \\ \hline
      \makecell{KG-Enhanced LLM~                                                                                       \\ \cite{gao2025leveraging}}                                          & \cmark                & \xmark     & \wmark         & \xmark           \\ \hline
      \makecell{AutoKG~                                                                                                \\ \cite{chen2023autokgefficientautomatedknowledge}}                       & \cmark                & \cmark          & \xmark           & \xmark           \\ \hline
      \makecell{\textbf{\sysname{} (Ours)}} & \textbf{\cmark}    & \textbf{\cmark} & \textbf{\cmark} & \textbf{\cmark} \\
      \bottomrule
    \end{tabular}
  }
  \vspace{-3mm}
\end{table}

As shown in Table~\ref{tab:rw-compare}, we categorize related work into three categories and compare them in Table~\ref{tab:rw-compare}.

\noindent \textbf{Sequential Diagnosis Agents.} Sequential diagnosis requires agents to balance information acquisition and diagnostic costs over multiple interactions. Recently, \citet{nori2025sequentialdiagnosislanguagemodels} proposed SDBench and MAI-DxO, achieving diagnostic accuracy and cost control surpassing human doctors through multi-agent orchestration. \citet{jia2025ddodualdecisionoptimizationllmbased} proposed the DDO framework, using Reinforcement Learning (RL) to optimize symptom inquiry strategies. \citet{schmidgall2025agentclinicmultimodalagentbenchmark} and \citet{qiu2025evolvingdiagnosticagentsvirtual} built AgentClinic and DiagGym simulation environments respectively for training and evaluating multimodal diagnostic agents. Additionally, ACTMED proposed by \citet{estévez2025timelyclinicaldiagnosisactive} combines Bayesian experimental design with LLMs to actively select tests. However, most of these methods rely on the implicit reasoning of LLMs or RL with low sample efficiency, lacking an explicit, interpretable structured domain model to guide decisions.

\noindent \textbf{Automated Medical Knowledge Graph Construction.} Traditional MKG construction relies on expert annotation and is costly. Recent research has begun to explore using LLMs for automated graph construction. \citet{chen2023autokgefficientautomatedknowledge} proposed AutoKG, using LLMs to extract keywords and calculate graph Laplacian weights. CoDe-KG by \citet{anuyah-etal-2025-automated} uses sentence complexity modeling to extract triples. \citet{sarabadani2025dkgllmframeworkmedical} proposed DKG-LLM, integrating dynamic knowledge graphs with the Grok 3 model. Although these works achieve automation, the constructed graphs are mostly used for semantic retrieval and lack the fine-grained typicality parameters (e.g., symptom weights) and decision utility attributes (e.g., test cost, invasiveness) required for diagnostic reasoning.

\noindent \textbf{Knowledge-Enhanced Clinical Decision.} Knowledge-Enhanced AI aims to combine the perceptual capabilities of neural networks with the logical capabilities of symbolic systems. \citet{gao2025leveraging} explored integrating MKG into LLMs for diagnostic prediction. Our method goes a step further, not only achieving automated graph construction but also designing an explicit evidence scoring engine based on the graph, realizing true "knowledge-guided reasoning."

In summary, our method \sysname{} is the only framework that simultaneously possesses automated graph construction capabilities, supports sequential diagnosis, and explicitly models test costs.

\section{Methodology}
\label{sec:methodology}

\subsection{Problem Formulation}
We formulate sequential diagnosis as a partially observable decision process. Let $\mathcal{D}$ be the set of possible diseases. At turn $t$, the agent observes dialogue history $H_t$ and maintains a belief state over $\mathcal{D}$. The agent selects an action $a_t \in \mathcal{A}_{action}$:
\begin{itemize}[leftmargin=*, nosep]
  \item \textbf{Inquire}$(f)$: Ask about symptom or feature $f$.
  \item \textbf{Test}$(\tau)$: Order diagnostic test $\tau$.
  \item \textbf{Diagnose}$(d)$: Provide final diagnosis $d$.
\end{itemize}
The objective is to minimize the expected total cost:
\begin{equation}
  \min_\pi \mathbb{E}\left[\sum_{t=1}^{T} c(a_t) + \lambda_{mis} \cdot \mathbb{I}[\hat{d} \neq d^*]\right]
\end{equation}
where $c(a_t)$ is the cost of action $a_t$ (including monetary cost, invasiveness, and time), $\lambda_{mis}$ is the penalty for misdiagnosis, $\hat{d}$ is the predicted diagnosis, and $d^*$ is the ground truth.

\subsection{Automated Construction of Medical Diagnosis Knowledge Graph}
\label{sec:mdkg_construction}
The core of \sysname{} is the Medical Diagnosis Knowledge Graph (MDKG), formally defined as a directed graph $\mathcal{G} = (\mathcal{V}, \mathcal{E}, \mathcal{W}, \mathcal{A})$ where:
\begin{itemize}[leftmargin=*, nosep]
  \item $\mathcal{V} = \mathcal{V}_d \cup \mathcal{V}_f \cup \mathcal{V}_t$: nodes for diseases, features (symptoms/signs), and tests.
  \item $\mathcal{E} \subseteq (\mathcal{V}_d \times \mathcal{V}_f) \cup (\mathcal{V}_f \times \mathcal{V}_t)$: edges connecting diseases to features and features to tests.
  \item $\mathcal{W}: \mathcal{E} \to \mathbb{R}$: edge weights encoding association strength.
  \item $\mathcal{A}: \mathcal{V}_t \to \mathbb{R}^2$: test attributes (cost, invasiveness).
\end{itemize}

We implement a fully automated pipeline to construct $\mathcal{G}$ from a disease list $\mathcal{D}_{list}$ without manual annotation. The process proceeds in three stages, designed to ensure the graph serves as a robust computational substrate for evidence-based reasoning.

\begin{description}[style=unboxed, leftmargin=0cm]
  \item[Stage 1: Parallel Knowledge Extraction.]
        The system processes each disease $d \in \mathcal{D}_{list}$ in parallel. For each disease, it prompts the LLM to extract a set of tuples $\{(f, w_{df}, T_f)\}$, where $f$ is a feature, $w_{df}$ is its typicality, and $T_f$ is a list of verifying tests.
        To mitigate the noise from poorly calibrated continuous probability estimates in LLMs, we employ a \textbf{Quantized Typicality} strategy. We map linguistic typicality descriptions to discrete weight buckets $\mathcal{W}: \mathcal{E}_{df} \to \mathbb{R}$:
        \begin{equation}
          w(d, f) = \begin{cases}
            w_{++} & \text{if } f \text{ is Hallmark (++})   \\
            w_{+}  & \text{if } f \text{ is Common (+)}      \\
            w_{-}  & \text{if } f \text{ is Rare (-)}        \\
            w_{--} & \text{if } f \text{ is Impossible (--)}
          \end{cases}
        \end{equation}
        This quantization acts as a low-pass filter, stabilizing the reasoning process against minor probability fluctuations in LLM generation.

        Additionally, to ensure the graph directly supports diagnostic planning, we enforce an \textbf{Action-Centric Topology}. Instead of just declarative knowledge (``Pneumonia causes cough''), the extraction explicitly models \emph{verification edges} $(f, \tau) \in \mathcal{E}_{f\tau}$ connecting features to the tests that can confirm or rule them out. This transforms the graph into a navigation structure: given an uncertain feature, the agent can directly traverse to relevant diagnostic actions.

  \item[Stage 2: Hybrid Entity Alignment \& Dynamic Upgrade.]
        Medical concepts often suffer from semantic ambiguity and role shifts (e.g., a disease in one context is a symptom in another). To resolve synonymy, we employ a \textbf{Hybrid Alignment Mechanism} to align a new term $u$ to the existing node set $\mathcal{V}$.
        We retrieve a candidate set $\mathcal{C}_u$ using a hybrid retrieval strategy that combines lexical and semantic matching:
        \begin{equation}
          \mathcal{C}_u = \text{TopK}_{\text{overlap}}(u, \mathcal{V}) \cup \text{TopK}_{\text{embed}}(u, \mathcal{V})
        \end{equation}
        where $\text{TopK}_{\text{overlap}}$ selects candidates based on the Overlap Coefficient (intersection over minimum set size) to capture partial matches (e.g., "Diabetes" $\subseteq$ "Type 2 Diabetes"), and $\text{TopK}_{\text{embed}}$ selects candidates based on cosine similarity of dense embeddings $\phi(\cdot)$ to capture semantic synonyms. An LLM then acts as a discriminator $D(u, \mathcal{C}_u)$ to determine if $u$ should merge with $v^* \in \mathcal{C}_u$ or form a new node.

        Furthermore, to handle role shifts without manual intervention, we introduce a \textbf{Dynamic Node Upgrade} strategy. When a node initially added as a feature (e.g., ``Heart Failure'' as a symptom of renal disease) is later identified as an independent disease, the system:
        \begin{enumerate}
          \item Promotes the node from $\mathcal{V}_f$ to $\mathcal{V}_d$: $\mathcal{V}_d \leftarrow \mathcal{V}_d \cup \{v\}, \mathcal{V}_f \leftarrow \mathcal{V}_f \setminus \{v\}$
          \item Preserves all existing incoming edges $(d', v)$ (its role as a feature)
          \item Enables new outgoing edges $(v, f')$(its features as a disease)
        \end{enumerate}

        This allows complex causal chains (e.g., Renal Disease $\to$ Heart Failure $\to$ Dyspnea) to emerge automatically without manual ontology restructuring.

  \item[Stage 3: Attribute Enrichment.]
        Finally, to support cost-sensitive decision making, the graph requires quantitative metadata often missing from raw extractions. The \textbf{Metadata Enrichment Module} infers these attributes for new test nodes. Each test node $\tau \in \mathcal{V}_t$ is associated with attributes $\mathcal{A}(\tau) = (c_\tau, \iota_\tau)$, where $c_\tau$ represents the monetary cost level and $\iota_\tau$ represents the invasiveness multiplier.
        This ensures the graph is fully populated with the quantitative data required for downstream planning.
\end{description}

\subsection{Graph-Augmented Diagnostic Agent}
\label{sec:reasoning}
To bridge the gap between unstructured patient interaction and structured medical reasoning, we design a \textbf{Graph-Augmented Diagnostic Agent}. This agent operates through a "Think-then-Act" loop (Algorithm 2), where a neural component handles perception and a symbolic component handles reasoning.

\subsubsection{Perception Agent}
Since patient inputs are unstructured and diverse (ranging from verbal descriptions to medical reports), the system must first normalize them into a structured format. The \textbf{Perception Module} acts as this interface. At turn $t$, it employs an extraction function $f_{ext}$ to parse the input $r_t$ (which may be a patient's verbal response or a returned medical test report) into structured observations:
\begin{equation}
  O_t = f_{ext}(r_t | H_{t-1}) = \{(m_i, s_i)\}_{i=1}^{K}
\end{equation}
where $m_i$ is the mention text (e.g., "fever" or "WBC count elevated") and $s_i \in \{\text{Present}, \text{Absent}\}$ is the status. Crucially, this unified extraction handles both subjective symptoms and objective test findings identically.
Subsequently, a grounding function $g: \mathcal{M} \to \mathcal{V}_f$ maps each mention $m_i$ to a canonical graph node $f^*_i$ using the alignment mechanism from Sec.~\ref{sec:mdkg_construction}. The symbolic state is updated as $\mathcal{S}_t = \mathcal{S}_{t-1} \cup \{(g(m_i), s_i)\}_{i=1}^K$.

\subsubsection{Reasoning Agent}
Once observations are grounded, the core challenge is to make deterministic and cost-effective decisions. The \textbf{Reasoning Engine} addresses this through two key mechanisms:

\paragraph{Evidence Scoring.}
To avoid the instability of black-box neural reasoning, we implement an explicit \textbf{Evidence Scoring} mechanism. It computes a confidence score $S(d)$ for each disease $d$ based on the alignment between the graph knowledge and the observed state $\mathcal{S}_t$. The scoring function explicitly handles positive and negative evidence:
\begin{equation}
  S(d) = \sum_{(f, s) \in \mathcal{S}_t} \text{Score}(f, d, s)
\end{equation}
where the component score is defined as:
\begin{equation}
  \text{Score}(f, d, s) = \begin{cases}
    w(d,f)                & \text{if } s=\text{Present} \\
    -\lambda \cdot w(d,f) & \text{if } s=\text{Absent}
  \end{cases}
\end{equation}
Here, $\lambda$ is a penalty factor. Note that if $w(d,f) < 0$ (the disease typically lacks the feature) and $s=\text{Present}$, the term $w(d,f)$ naturally reduces the score. Conversely, if $s=\text{Absent}$ and $w(d,f) > 0$ (expected feature missing), the term $-\lambda \cdot w(d,f)$ penalizes the disease.

\paragraph{Cost-Sensitive Action Planning.}
To balance diagnostic accuracy with cost (mitigating "defensive medicine"), we employ a \textbf{Cost-Sensitive Action Planning} module. While full Bayesian Value of Information (VoI) calculation is computationally prohibitive for real-time interaction, we implement a \emph{heuristic utility function} that approximates information gain via feature variance and relevance:

\begin{itemize}[leftmargin=*]
  \item \textbf{Feature Selection.} We identify unobserved features that maximize discriminative power among the top-$K$ candidate diseases $D_{top}$. The discriminative score $V(f)$ serves as a lightweight proxy for information gain:
        \begin{equation}
          \begin{split}
            V(f) & = \sum_{d \in D_{top}} \lvert S(d) \rvert \cdot \lvert w(d,f) \rvert \\
                 & \quad + \alpha \cdot \operatorname{Var}_{d \in D_{top}}(w(d,f))
          \end{split}
        \end{equation}
        The first term prioritizes features relevant to high-confidence diseases, while the second term (variance) favors features that can differentiate between them.

  \item \textbf{Test Selection.} For each candidate test $\tau$, we calculate its utility $U(\tau)$ by aggregating the value of the features it can verify:
        \begin{equation}
          U(\tau) = \frac{\sum_{f \in \text{Verifies}(\tau)} V(f) \cdot \eta(f,\tau)}{1 + c_\tau \cdot \iota_\tau}
        \end{equation}
        where $\eta(f,\tau) \in \{\eta_{low}, \eta_{high}\}$ denotes the effectiveness of test $\tau$ for feature $f$. The denominator penalizes expensive and invasive tests, ensuring the selected actions maximize the information-to-cost ratio.
\end{itemize}

\subsubsection{Decision Agent}
Acting as the "right brain," this agent generates the final action $a_t$ and natural language response $resp_t$ by conditioning on the structured reasoning summary $\Sigma_t$:
\begin{equation}
  (a_t, resp_t) \sim \pi_{\theta}( \cdot \mid H_t, \Sigma_t)
\end{equation}
where $\Sigma_t$ aggregates the symbolic outputs:
\begin{equation}
  \Sigma_t = \left\langle \text{Rank}(\mathcal{D}), \text{Top}_K^{V}(f), \text{Top}_K^{U}(\tau) \right\rangle
\end{equation}
This injection of $\Sigma_t$ constrains the LLM's generation space to medically valid paths derived from the MDKG.

\subsection{Design Rationale}
The \sysname{} architecture embodies a key insight: \emph{neural and symbolic components excel at complementary tasks}. LLMs handle the ``soft'' aspects of diagnosis---understanding colloquial language, maintaining conversational flow, and exercising clinical judgment in ambiguous situations. The symbolic engine handles the ``hard'' aspects---deterministic state tracking, systematic hypothesis scoring, and principled cost-benefit analysis. This separation provides several advantages: (1) \textbf{Interpretability}: Every diagnostic conclusion traces to explicit evidence scores; (2) \textbf{Consistency}: Deterministic state tracking prevents contradictory reasoning; (3) \textbf{Efficiency}: Symbolic utility computation is faster than LLM deliberation; (4) \textbf{Robustness}: Graph structure constrains LLM outputs to valid medical reasoning.

\section{Experiments}
\label{sec:experiments}
We conduct extensive experiments to address the following research questions: \textbf{Q1}: Can \sysname{} achieve superior diagnostic accuracy and cost-efficiency compared to current SOTA LLM frameworks? \textbf{Q2}: How robust is our method when transferring to real-world, noisy clinical scenarios? \textbf{Q3}: What are the individual contributions of the structured MDKG and the explicit inference engine?

\paragraph{Implementation Details.} We evaluate \sysname{} on two datasets: \textbf{MedQA} (200 cases) and \textbf{MIMIC-IV} (200 real-world cases). We employ \texttt{ClinicSim}, a modular simulation environment, to conduct multi-turn diagnostic dialogues. We compare our method against \textbf{Standard LLM} prompting and the multi-agent \textbf{MAI-DxO} framework~\cite{nori2025sequentialdiagnosislanguagemodels} across three LLM backbones: DeepSeek-V3, Kimi-k2, and Llama-3.3. Performance is measured by Diagnostic Score (0-10), Total Cost (\$), and Dialogue Turns. For detailed descriptions of the datasets, baseline settings, hyperparameter configurations, and the simulation environment, please refer to Appendix~\ref{app:detailed_setup}.

\subsection{Experimental Results}

\subsubsection{Main Results}
Table~\ref{tab:main_results} and Table~\ref{tab:mimiciv_results} present the performance comparison between \sysname{} and baselines on MedQA and MIMIC-IV datasets, respectively. The results demonstrate that our method achieves consistent and significant improvements across all metrics and base models.

\paragraph{Diagnostic Accuracy.}
On the MedQA dataset, \sysname{} significantly outperforms the Standard LLM baseline. For instance, with DeepSeek-V3, the success rate (Score $\ge$ 8) jumps from 67.8\% to 88.7\%, and Kimi-k2 achieves a remarkable 93.2\% success rate (assessment details in Appendix~\ref{app:eval_prompts}).
We also compared our method with the SOTA multi-agent framework MAI-DxO~\cite{nori2025sequentialdiagnosislanguagemodels}. While MAI-DxO improves accuracy on DeepSeek-V3 (8.05), it incurs extremely high costs (\$2460) and interaction turns (9.51). On the smaller Llama-3.3 model, MAI-DxO suffers from performance collapse (Score 5.38), indicating its high dependency on the base model's capability. In contrast, \sysname{} achieves Pareto optimality across all models by offloading reasoning to the graph.
This advantage generalizes well to the real-world MIMIC-IV dataset. Despite the increased clinical noise, \sysname{} improves the average diagnostic score by approximately 2.0 points (e.g., DeepSeek-V3: 6.60 $\to$ 8.27) and increases the success rate by 25\%--30\%. In contrast, the MAI-DxO baseline struggles significantly on this real-world dataset, showing lower scores and high variance, likely due to the complexity of real clinical notes confusing the multi-agent debate mechanism. This proves that structured knowledge effectively guides the model to grasp key features even in noisy environments.

\begin{figure}[t!]
  \centering
  \includegraphics[width=\linewidth]{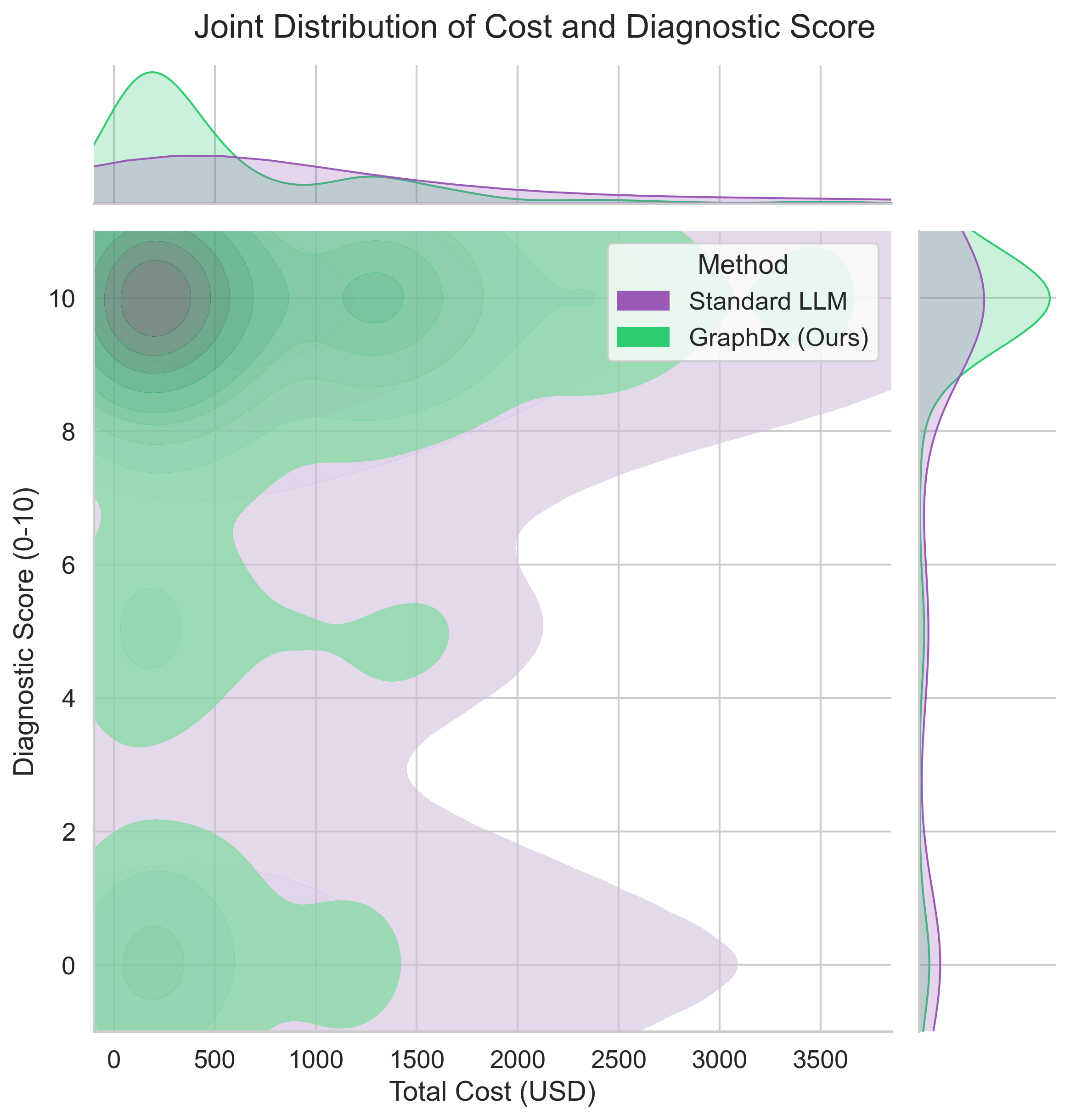}
  \vspace{-6mm}
  \caption{Joint distribution of Cost vs. Diagnostic Score for DeepSeek-V3 model.}
  \label{fig:cost_acc_scatter}
  \vspace{-5mm}
\end{figure}

\paragraph{Cost-Effectiveness and Efficiency.}
\sysname{} consistently reduces test costs by 20\%--54\% across both datasets while improving accuracy (see Appendix~\ref{sec:cost_estimation} for details on the US-based cost estimation methodology). For DeepSeek-V3 on MedQA, the average cost drops from \$1062 to \$537. This is attributed to the utility-based active planning module, which effectively suppresses the "defensive medicine" behavior—blindly ordering expensive tests (e.g., full-body CT) to compensate for uncertainty—observed in baselines. Notably, on MIMIC-IV, MAI-DxO incurs extremely high costs (e.g., \$1497 for DeepSeek-V3) due to uncontrolled exploration, whereas \sysname{} maintains low costs (\$340) by precisely targeting relevant tests.
Figure~\ref{fig:cost_acc_scatter} illustrates this advantage: while Baseline results (purple) are scattered with many high-cost, low-score samples, \sysname{} (green) is tightly clustered in the "high score-low cost" region.
Regarding interaction efficiency, \sysname{} incurs slightly more turns than the Standard LLM (e.g., 5.06 vs. 3.80 for Llama-3.3). This is deliberate: the graph agent performs more thorough differential diagnosis, excluding distractors through multi-turn inquiries rather than jumping to premature conclusions. Given the substantial gains in diagnostic quality, this trade-off is clinically justified.

\paragraph{Robustness and Failure Analysis.}
On the challenging MIMIC-IV dataset, we observed that baselines often fail due to (1) \textbf{Noise Interference}, being misled by irrelevant details in past medical history (e.g., old fractures), and (2) \textbf{Defensive Medicine}, leading to soaring costs (e.g., Scenario 66 incurred \$1825 in costs yet still resulted in misdiagnosis). \sysname{} mitigates these issues by filtering noise through the MDKG's prior knowledge, focusing reasoning on core pathological features and avoiding unnecessary expensive testing.

\begin{table*}[t]
  \centering
  \caption{Diagnostic Performance Comparison on MedQA Dataset}
  \label{tab:main_results}
  \resizebox{\linewidth}{!}{
    \begin{tabular}{llccccc}
      \toprule
      \textbf{Base Model}          & \textbf{Method}            & \textbf{Score (0-10)} $\uparrow$ & \textbf{Cost (\$)} $\downarrow$ & \textbf{Turns}   & \textbf{Success Rate} $\uparrow$ & \textbf{Harmful Rate} $\downarrow$ \\
      \midrule
      \multirow{3}{*}{DeepSeek-V3} & Standard LLM               & 7.25 $\pm$ 1.34                  & 1062.69 $\pm$ 655.20            & 4.66 $\pm$ 0.77  & 67.8\%                           & 22.5\%                             \\
                                   & MAI-DxO                    & 8.05 $\pm$ 0.98                  & 2460.20 $\pm$ 1114.69           & 9.51 $\pm$ 2.74  & 75.0\%                           & 13.2\%                             \\
                                   & \textbf{\sysname{} (Ours)} & \textbf{9.05 $\pm$ 0.47}         & \textbf{537.13 $\pm$ 221.24}    & 4.76 $\pm$ 0.94  & \textbf{88.7\%}                  & \textbf{7.7\%}                     \\
      \midrule
      \multirow{3}{*}{Kimi-k2}     & Standard LLM               & 7.66 $\pm$ 1.24                  & 1456.88 $\pm$ 575.35            & 3.62 $\pm$ 0.51  & 66.8\%                           & 12.2\%                             \\
                                   & MAI-DxO                    & 6.92 $\pm$ 1.61                  & 2220.57 $\pm$ 1026.20           & 11.99 $\pm$ 3.54 & 64.0\%                           & 24.5\%                             \\
                                   & \textbf{\sysname{} (Ours)} & \textbf{9.50 $\pm$ 0.29}         & \textbf{1156.02 $\pm$ 407.83}   & 4.17 $\pm$ 0.73  & \textbf{93.2\%}                  & \textbf{1.3\%}                     \\
      \midrule
      \multirow{3}{*}{Llama-3.3}   & Standard LLM               & 7.51 $\pm$ 1.30                  & 1857.47 $\pm$ 773.05            & 3.80 $\pm$ 0.68  & 65.7\%                           & 12.5\%                             \\
                                   & MAI-DxO                    & 5.38 $\pm$ 2.06                  & 3029.19 $\pm$ 1236.24           & 16.86 $\pm$ 2.01 & 50.3\%                           & 42.3\%                             \\
                                   & \textbf{\sysname{} (Ours)} & \textbf{9.24 $\pm$ 0.57}         & \textbf{1065.78 $\pm$ 469.34}   & 5.06 $\pm$ 1.51  & \textbf{88.3\%}                  & \textbf{4.3\%}                     \\
      \bottomrule
    \end{tabular}
  }
  \vspace{-2mm}
\end{table*}

\begin{table*}[t]
  \centering
  \caption{Diagnostic Performance Comparison on MIMIC-IV Dataset}
  \label{tab:mimiciv_results}
  \resizebox{\linewidth}{!}{
    \begin{tabular}{llccccc}
      \toprule
      \textbf{Base Model}          & \textbf{Method}            & \textbf{Score (0-10)} $\uparrow$ & \textbf{Cost (\$)} $\downarrow$ & \textbf{Turns}   & \textbf{Success Rate} $\uparrow$ & \textbf{Harmful Rate} $\downarrow$ \\
      \midrule
      \multirow{3}{*}{DeepSeek-V3} & Standard LLM               & 6.60 $\pm$ 0.93                  & 737.43 $\pm$ 267.38             & 5.41 $\pm$ 0.87  & 56.8\%                           & 18.8\%                             \\
                                   & MAI-DxO                    & 5.57 $\pm$ 1.43                  & 1497.13 $\pm$ 844.00            & 11.28 $\pm$ 3.53 & 45.0\%                           & 30.5\%                             \\
                                   & \textbf{\sysname{} (Ours)} & \textbf{8.27 $\pm$ 0.51}         & \textbf{340.98 $\pm$ 129.84}    & 5.19 $\pm$ 0.93  & \textbf{83.0\%}                  & \textbf{5.7\%}                     \\
      \midrule
      \multirow{3}{*}{Kimi-k2}     & Standard LLM               & 6.24 $\pm$ 1.08                  & 1040.60 $\pm$ 539.29            & 4.02 $\pm$ 0.57  & 54.2\%                           & 23.2\%                             \\
                                   & MAI-DxO                    & 5.19 $\pm$ 1.45                  & 956.93 $\pm$ 559.68             & 11.97 $\pm$ 3.71 & 45.0\%                           & 36.5\%                             \\
                                   & \textbf{\sysname{} (Ours)} & \textbf{8.06 $\pm$ 0.65}         & \textbf{658.48 $\pm$ 246.00}    & 5.30 $\pm$ 1.26  & \textbf{79.3\%}                  & \textbf{7.5\%}                     \\
      \midrule
      \multirow{3}{*}{Llama-3.3}   & Standard LLM               & 6.14 $\pm$ 1.00                  & 1164.27 $\pm$ 521.71            & 4.34 $\pm$ 0.66  & 50.3\%                           & 22.0\%                             \\
                                   & MAI-DxO                    & 3.56 $\pm$ 1.43                  & 1820.21 $\pm$ 803.68            & 17.53 $\pm$ 1.65 & 30.8\%                           & 57.2\%                             \\
                                   & \textbf{\sysname{} (Ours)} & \textbf{8.55 $\pm$ 0.67}         & \textbf{711.93 $\pm$ 331.54}    & 6.60 $\pm$ 1.85  & \textbf{82.2\%}                  & \textbf{5.7\%}                     \\
      \bottomrule
    \end{tabular}
    \vspace{-2mm}
  }
\end{table*}

\subsubsection{Performance on Unseen Diseases (Open-Set Evaluation)}
\label{sec:openset}
To further address the concern regarding the closed-set nature of the MDKG, we conducted an "Open-Set" evaluation using a subset of MIMIC-IV cases where the ground truth disease was \textbf{explicitly excluded} from our constructed graph. This setting tests the system's ability to handle "unknown" conditions by relying on the hybrid knowledge-enhanced architecture.

Results on the DeepSeek-V3 backbone show that even without specific disease nodes, \sysname{} outperforms the Standard LLM baseline. \sysname{} achieved a mean score of \textbf{7.65} compared to the baseline's \textbf{6.66}, a significant improvement of \textbf{+1.0 point}. The average test cost was reduced from \textbf{\$743.48} (Baseline) to \textbf{\$574.88} (\sysname{}), a saving of \textbf{$\sim$23\%}.

These findings confirm the robustness of our hybrid design. Although the specific disease node is missing, the "Unaligned Features" mechanism (Algorithm~\ref{alg:inference}) ensures that clinical observations are still captured and passed to the LLM. Furthermore, the graph's general structure of feature-test relationships remains valid (e.g., "Chest Pain" suggests "ECG" regardless of the specific underlying pathology), allowing the agent to conduct efficient workups before falling back on the LLM's parametric knowledge for the final diagnosis.

\subsubsection{Human Evaluation of Simulation Quality}
To address concerns regarding the reliance on simulation and LLM-based evaluation, we conducted a rigorous validation study to ensure clinical validity. Medical experts reviewed 365 diagnostic scenarios and 143 medical test cost estimates generated by our system. We analyzed the consistency between human experts and our automated LLM judge. The Pearson correlation coefficient for \textbf{Diagnostic Score} was \textbf{0.89} (n=365), with a Mean Absolute Error (MAE) of \textbf{0.58} (on a 0-10 scale) and a \textbf{94.8\%} agreement rate on success/failure classification (Score $\ge$ 8). For \textbf{Test Cost}, the correlation was \textbf{0.89} (n=143) with an MAE of \textbf{\$226.14}. These results provide strong empirical evidence that our automated evaluation pipeline is a reliable proxy for human expert judgment.

% These results provide strong empirical evidence that our automated evaluation pipeline is a reliable proxy for human expert judgment, effectively mitigating the validity concerns often associated with LLM-as-a-Judge methodologies. Furthermore, it is important to note that the "Ground Truth" diagnoses used for evaluation are derived directly from the gold-standard labels of the MedQA and MIMIC-IV datasets, not generated by the simulator, ensuring that the evaluation target remains objective and clinically valid.

% \subsubsection{Interaction Efficiency Analysis}
% Regarding interaction turns (Table~\ref{tab:main_results}), the number of turns for \sysname{} is slightly higher than Baseline (e.g., Llama-3.3 increased from 3.80 to 5.06). This is because the graph agent tends to perform more detailed differential diagnosis, excluding distractors through multi-turn inquiries rather than jumping to conclusions. Given the huge improvement in diagnostic quality, this slight cost in efficiency is completely acceptable. Detailed analysis is provided in Appendix~\ref{app:interaction_efficiency}.

\subsubsection{Visualization of Reasoning Process}
\begin{figure}[htbp]
  \centering
  \includegraphics[width=\linewidth]{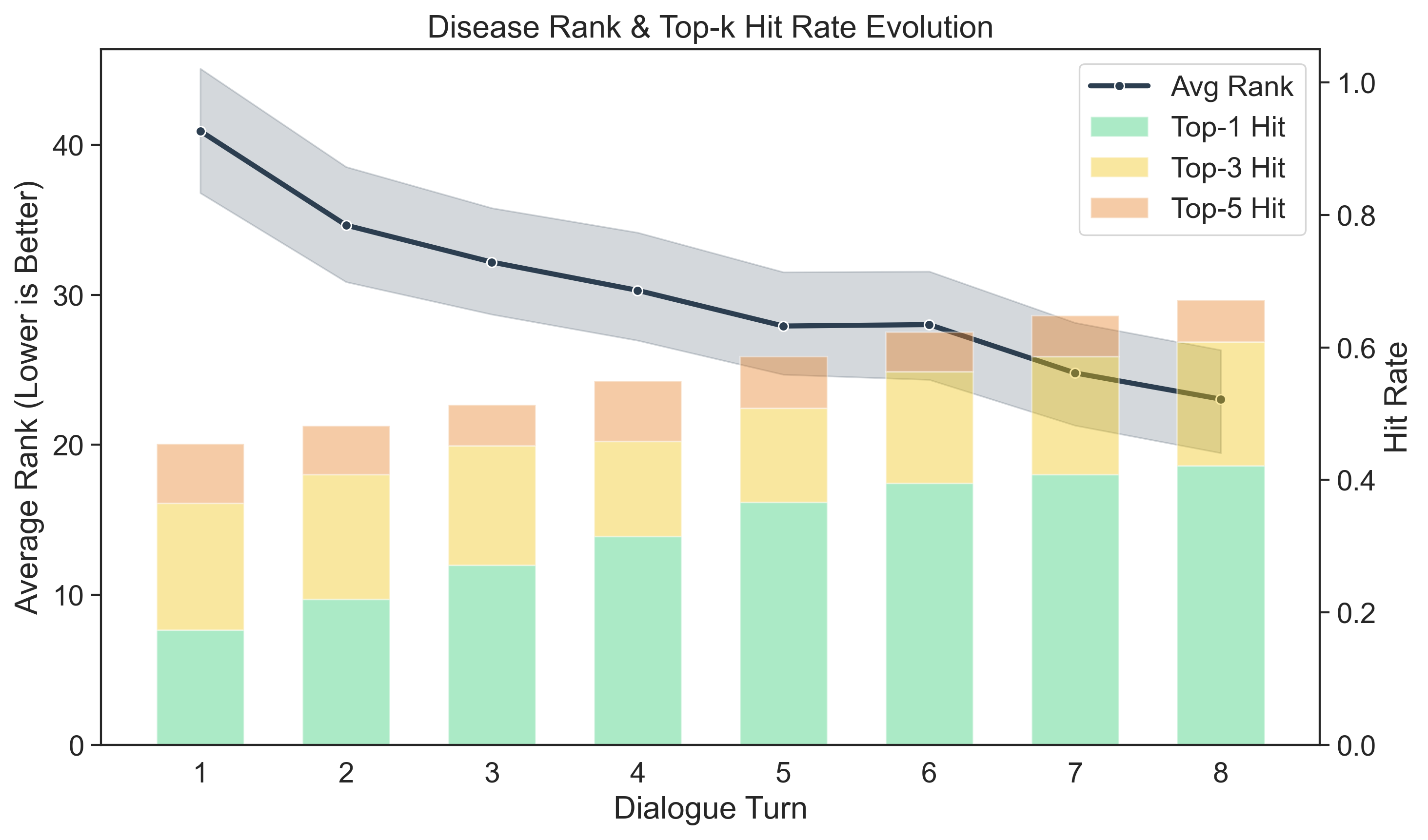}
  \vspace{-6mm}
  \caption{Evolution of GT Disease Rank and Top-k Hit Rate. The blue line (left) shows the decline in average rank, while bars (right) show the increasing probability of GT entering the Top-1/3/5 list.}
  \label{fig:rank_evolution}
  \vspace{-5mm}
\end{figure}

Figure~\ref{fig:rank_evolution} illustrates the evolution of the Ground Truth (GT) disease rank and Top-k hit rate over dialogue turns using DeepSeek-V3. We observe two key trends: (1) \textbf{Rapid Convergence:} In early turns (1-3), the GT rank drops sharply and Top-5 hit rate exceeds 50\%, showing the graph's ability to quickly lock onto relevant diseases. (2) \textbf{Effective Differential Diagnosis:} In later turns (4-8), the Top-1 hit rate rises above 40\%, confirming that our utility-based strategy effectively pinpoints the correct diagnosis through discriminative feature collection.

\subsubsection{Ablation Study}
\label{sec:ablation}
To quantify the contribution of each component (MDKG knowledge base and inference engine) in \sysname{}, we conducted an ablation study on the Llama-3.3 model using the MedQA-Extended dataset. We designed two variants:
\begin{itemize}[leftmargin=*, nosep]
  \item \textbf{w/o Graph:} Removes the real MDKG and instead uses the LLM to "hallucinate" a graph summary based on dialogue history. This tests the value of structured knowledge itself.
  \item \textbf{w/o Inference:} Retains the real MDKG for retrieving neighbor nodes but removes evidence scoring and utility-based planning, relying entirely on the LLM for decision-making. This tests the value of the explicit inference engine.
\end{itemize}

The experimental results are shown in Table~\ref{tab:ablation_data}.

\begin{table}[htbp]
  \centering
  \caption{Detailed Ablation Study Results on Llama-3.3}
  \label{tab:ablation_data}
  \resizebox{\linewidth}{!}{
    \begin{tabular}{lcccc}
      \toprule
      \textbf{Method}         & \textbf{Score} $\uparrow$ & \textbf{Cost (\$)} $\downarrow$ & \textbf{\makecell{Success                  \\ Rate}} $\uparrow$ & \textbf{\makecell{Harmful                  \\ Rate}} $\downarrow$ \\
      \midrule
      Baseline                & 7.51 $\pm$ 1.30           & 1857 $\pm$ 773                  & 65.7\%                    & 12.5\%         \\
      w/o Graph               & 8.00 $\pm$ 1.21           & 1710 $\pm$ 739                  & 71.5\%                    & 10.5\%         \\
      w/o Inference           & 8.39 $\pm$ 1.08           & 1286 $\pm$ 642                  & 77.5\%                    & 11.2\%         \\
      \textbf{GraphDx (Full)} & \textbf{9.24 $\pm$ 0.57}  & \textbf{1066 $\pm$ 469}         & \textbf{88.3\%}           & \textbf{4.3\%} \\
      \bottomrule
    \end{tabular}
  }
\end{table}

The analysis is as follows:
\begin{enumerate}[leftmargin=*]
  \item \textbf{Necessity of Structured Knowledge:} Compared to Baseline, \textbf{w/o Graph} shows a slight performance improvement (Score 7.51 $\to$ 8.00), indicating that even a hallucinated structured summary helps the LLM organize its thoughts. However, \textbf{w/o Inference} (using the real graph) further improves performance (Score $\to$ 8.39, Cost $\downarrow$ 25\%), proving that accurate domain knowledge (MDKG) is far superior to the LLM's parametric memory.
  \item \textbf{Critical Role of Explicit Reasoning:} The full \sysname{} achieves the largest performance leap compared to \textbf{w/o Inference} (Score $\to$ 9.24, Cost $\downarrow$ 17\%). This demonstrates that knowledge alone is not enough; the explicit inference engine based on evidence scoring and utility effectively guides the agent to make optimal decisions in complex uncertain environments, avoiding the LLM's blind trial-and-error.
\end{enumerate}

\subsubsection{Parameter Sensitivity Analysis}
\label{sec:sensitivity}
To assess the robustness of our mechanism, we conducted an offline sensitivity analysis on the interaction logs. We re-evaluated the disease ranking logic by varying key hyperparameters. The results, detailed in Appendix~\ref{app:sensitivity}, show that the system maintains stable performance across a wide range of parameters, demonstrating robustness.

\subsubsection{Case Study}
To deeply understand the decision-making advantages of \sysname{} in different clinical contexts, we selected representative cases for detailed comparative analysis. These cases reveal typical failure modes of baseline models in \textbf{rare disease reasoning}, \textbf{anatomical logic}, and \textbf{cost-effective diagnosis}, while \sysname{} successfully avoids these pitfalls through structured knowledge. Detailed analysis and visualization are provided in Appendix~\ref{app:case_study}.

\section{Conclusion}
\label{sec:conclusion}
This paper proposes \sysname{}, which effectively solves the "Knowledge-Reasoning Mismatch" problem of LLMs in sequential diagnosis by automatically constructing a medical knowledge graph and combining it with knowledge-enhanced reasoning. Experiments demonstrate that this method not only significantly improves the accuracy and economy of diagnosis on standard medical exam datasets but also exhibits excellent generalization and robustness on real-world MIMIC-IV clinical data, providing a feasible path for the interpretability of medical AI. Future work will focus on extending this framework to multi-modal data fusion and larger-scale clinical deployment validation.

\section*{Limitations}
Despite the excellent performance of \sysname{} in experiments, there are still some limitations:
(1) \textbf{Graph Construction Depends on Base Model Capability}: The quality of MDKG is limited by the medical knowledge reserve of the LLM used for graph construction (DeepSeek-V3 in this paper). If the base model has knowledge blind spots, the graph may miss key nodes or edges.
(2) \textbf{Limited Multi-modal Processing Capability}: The current framework mainly processes textual information. The understanding of medical imaging (CT/MRI) and waveform data (ECG) still relies on the text descriptions from LLMs, and end-to-end pixel-level feature extraction has not yet been achieved.
(3) \textbf{Inference Speed}: Due to the introduction of explicit evidence scoring and utility calculation, the single-step decision latency of \sysname{} is slightly higher than that of pure end-to-end LLMs. Future work needs to further optimize the computational efficiency of the inference engine.
(4) \textbf{Simulation Gap}: While our simulator is designed to be realistic, simulated patients may exhibit more consistent logic than real patients, who might provide vague or misleading descriptions.
(5) \textbf{LLM-based Evaluation Bias}: Our simulator, judge, and cost estimator are LLM-based components and may inherit biases (e.g., stylistic preferences or calibration drift). While we mitigate this via a fixed evaluation protocol and human validation (Section~\ref{sec:experiments}), future work should incorporate multi-judge ensembles and real-world clinical evaluation when feasible.
\FloatBarrier

\section*{Ethics Statement}
This research involves the development of an automated medical diagnosis system. We emphasize that \sysname{} is intended for research and educational simulation purposes only and must not be used for direct clinical decision-making without human oversight. The datasets used (MedQA and MIMIC-IV) are de-identified public datasets, ensuring patient privacy. While the automated graph construction reduces manual effort, it may inherit biases or outdated knowledge from the base LLM; therefore, any deployment in real-world settings requires rigorous human-in-the-loop verification. Furthermore, while our system optimizes for cost-efficiency, we acknowledge the ethical trade-off between reducing costs and the potential risk of missing necessary tests, which warrants further investigation.

\bibliography{custom}

@inproceedings{anuyah-etal-2025-automated,
  title     = {Automated Knowledge Graph Construction using Large Language Models and Sentence Complexity Modelling},
  author    = {Anuyah, Sydney  and
               Kaushik, Mehedi Mahmud  and
               Dwarampudi, Sri Rama Krishna Reddy  and
               Shiradkar, Rakesh  and
               Durresi, Arjan  and
               Chakraborty, Sunandan},
  editor    = {Christodoulopoulos, Christos  and
               Chakraborty, Tanmoy  and
               Rose, Carolyn  and
               Peng, Violet},
  booktitle = {Proceedings of the 2025 Conference on Empirical Methods in Natural Language Processing},
  month     = nov,
  year      = {2025},
  address   = {Suzhou, China},
  publisher = {Association for Computational Linguistics},
  url       = {https://aclanthology.org/2025.emnlp-main.783/},
  doi       = {10.18653/v1/2025.emnlp-main.783},
  pages     = {15526--15550},
  isbn      = {979-8-89176-332-6}
}

@misc{chen2023autokgefficientautomatedknowledge,
  title         = {AutoKG: Efficient Automated Knowledge Graph Generation for Language Models},
  author        = {Bohan Chen and Andrea L. Bertozzi},
  year          = {2023},
  eprint        = {2311.14740},
  archiveprefix = {arXiv},
  primaryclass  = {cs.CL},
  url           = {https://arxiv.org/abs/2311.14740}
}

@misc{estévez2025timelyclinicaldiagnosisactive,
  title         = {Timely Clinical Diagnosis through Active Test Selection},
  author        = {Silas Ruhrberg Estévez and Nicolás Astorga and Mihaela van der Schaar},
  year          = {2025},
  eprint        = {2510.18988},
  archiveprefix = {arXiv},
  primaryclass  = {cs.AI},
  url           = {https://arxiv.org/abs/2510.18988}
}

@article{gao2025leveraging,
  title   = {Leveraging Medical Knowledge Graphs Into Large Language Models for Diagnosis Prediction: Design and Application Study},
  author  = {Gao, Y. and Li, R. and Croxford, E. and Caskey, J. and Patterson, B. and Churpek, M. and Miller, T. and Dligach, D. and Afshar, M.},
  journal = {JMIR AI},
  year    = {2025},
  volume  = {4},
  pages   = {e58670},
  url     = {https://ai.jmir.org/2025/1/e58670},
  doi     = {10.2196/58670}
}

@misc{jia2025ddodualdecisionoptimizationllmbased,
  title         = {DDO: Dual-Decision Optimization for LLM-Based Medical Consultation via Multi-Agent Collaboration},
  author        = {Zhihao Jia and Mingyi Jia and Junwen Duan and Jianxin Wang},
  year          = {2025},
  eprint        = {2505.18630},
  archiveprefix = {arXiv},
  primaryclass  = {cs.CL},
  url           = {https://arxiv.org/abs/2505.18630}
}

@misc{nori2025sequentialdiagnosislanguagemodels,
  title         = {Sequential Diagnosis with Language Models},
  author        = {Harsha Nori and Mayank Daswani and Christopher Kelly and Scott Lundberg and Marco Tulio Ribeiro and Marc Wilson and Xiaoxuan Liu and Viknesh Sounderajah and Jonathan Carlson and Matthew P Lungren and Bay Gross and Peter Hames and Mustafa Suleyman and Dominic King and Eric Horvitz},
  year          = {2025},
  eprint        = {2506.22405},
  archiveprefix = {arXiv},
  primaryclass  = {cs.CL},
  url           = {https://arxiv.org/abs/2506.22405}
}

@misc{qiu2025evolvingdiagnosticagentsvirtual,
  title         = {Evolving Diagnostic Agents in a Virtual Clinical Environment},
  author        = {Pengcheng Qiu and Chaoyi Wu and Junwei Liu and Qiaoyu Zheng and Yusheng Liao and Haowen Wang and Yun Yue and Qianrui Fan and Shuai Zhen and Jian Wang and Jinjie Gu and Yanfeng Wang and Ya Zhang and Weidi Xie},
  year          = {2025},
  eprint        = {2510.24654},
  archiveprefix = {arXiv},
  primaryclass  = {cs.CL},
  url           = {https://arxiv.org/abs/2510.24654}
}

@misc{sarabadani2025dkgllmframeworkmedical,
  title         = {DKG-LLM : A Framework for Medical Diagnosis and Personalized Treatment Recommendations via Dynamic Knowledge Graph and Large Language Model Integration},
  author        = {Ali Sarabadani and Maryam Abdollahi Shamami and Hamidreza Sadeghsalehi and Borhan Asadi and Saba Hesaraki},
  year          = {2025},
  eprint        = {2508.06186},
  archiveprefix = {arXiv},
  primaryclass  = {cs.CL},
  url           = {https://arxiv.org/abs/2508.06186}
}

@misc{schmidgall2025agentclinicmultimodalagentbenchmark,
  title         = {AgentClinic: a multimodal agent benchmark to evaluate AI in simulated clinical environments},
  author        = {Samuel Schmidgall and Rojin Ziaei and Carl Harris and Eduardo Reis and Jeffrey Jopling and Michael Moor},
  year          = {2025},
  eprint        = {2405.07960},
  archiveprefix = {arXiv},
  primaryclass  = {cs.HC},
  url           = {https://arxiv.org/abs/2405.07960}
}

@misc{nori2023capabilitiesgpt4medicalchallenge,
  title         = {Capabilities of GPT-4 on Medical Challenge Problems},
  author        = {Harsha Nori and Nicholas King and Scott Mayer McKinney and Dean Carignan and Eric Horvitz},
  year          = {2023},
  eprint        = {2303.13375},
  archiveprefix = {arXiv},
  primaryclass  = {cs.CL},
  url           = {https://arxiv.org/abs/2303.13375}
}

@article{singhalLargeLanguageModels2023,
  title    = {Large Language Models Encode Clinical Knowledge},
  author   = {Singhal, Karan and Azizi, Shekoofeh and Tu, Tao and Mahdavi, S. Sara and Wei, Jason and Chung, Hyung Won and Scales, Nathan and Tanwani, Ajay and {Cole-Lewis}, Heather and Pfohl, Stephen and Payne, Perry and Seneviratne, Martin and Gamble, Paul and Kelly, Chris and Babiker, Abubakr and Sch{\"a}rli, Nathanael and Chowdhery, Aakanksha and Mansfield, Philip and {Demner-Fushman}, Dina and {Ag{\"u}era y Arcas}, Blaise and Webster, Dale and Corrado, Greg S. and Matias, Yossi and Chou, Katherine and Gottweis, Juraj and Tomasev, Nenad and Liu, Yun and Rajkomar, Alvin and Barral, Joelle and Semturs, Christopher and Karthikesalingam, Alan and Natarajan, Vivek},
  year     = 2023,
  month    = aug,
  journal  = {Nature},
  volume   = {620},
  number   = {7972},
  pages    = {172--180},
  issn     = {1476-4687},
  doi      = {10.1038/s41586-023-06291-2}
}

@article{johnsonMIMICIVFreelyAccessible2023,
  title    = {{{MIMIC-IV}}, a Freely Accessible Electronic Health Record Dataset},
  author   = {Johnson, Alistair E. W. and Bulgarelli, Lucas and Shen, Lu and Gayles, Alvin and Shammout, Ayad and Horng, Steven and Pollard, Tom J. and Hao, Sicheng and Moody, Benjamin and Gow, Brian and Lehman, Li-wei H. and Celi, Leo A. and Mark, Roger G.},
  year     = 2023,
  month    = jan,
  journal  = {Scientific Data},
  volume   = {10},
  number   = {1},
  pages    = {1},
  issn     = {2052-4463},
  doi      = {10.1038/s41597-022-01899-x}
}

\appendix

\section{Prompts for Automated Graph Construction}
\label{app:prompts_construction}

We provide the key system prompts used in the MDKG construction pipeline. These prompts are designed to guide the LLM in extracting structured medical knowledge from its internal parameters.

\subsection{Disease Knowledge Extraction}
\label{app:prompt_disease}
This prompt instructs the LLM to extract key diagnostic features (symptoms, signs, risk factors) for a given disease and estimate their typicality.

\noindent\fbox{%
  \parbox{0.98\linewidth}{%
    \small
    \textbf{System Prompt:} \\
    You are a physician building a diagnostic knowledge graph.

    Input: ONE disease name (and optional reference text).
    Output: ONE JSON object with:
    - "disease": canonical name of this disease.
    - "features": an array of feature objects:
    - "name": short, canonical, self-explanatory term. No extra adjectives, no explanations. E.g. "fever", "chest pain", "elevated troponin", "chest X-ray consolidation".
    - "category": one of: "symptom", "test\_result", "risk\_factor", "other".
    - "typicality": one of "++", "+", "-", "--", meaning P(F|D):
    * "++": almost all patients with this disease have this feature (P(F|D) $\approx$ 1). Missing it makes the disease very atypical.
    * "+":  often present in this disease (P(F|D) clearly higher than baseline).
    * "-":  often absent in this disease (P(F|D) clearly lower than baseline).
    * "--": almost never present (P(F|D) $\approx$ 0). Seeing it strongly suggests other diseases.
    Important: "typicality" means frequency in this disease, NOT severity.
    - "tests": list of test names that DIRECTLY help observe or confirm THIS FEATURE, not "all tests for this disease". If none, use [].

    Rules:
    - Focus on important diagnostic features, not every trivial detail.
    - Prefer guideline/textbook-style canonical terminology.
    - You may use your medical knowledge beyond the given text but stay consistent.

    Return ONLY the JSON, no extra text.
  }%
}

\subsection{Term Alignment}
\label{app:prompt_align}
This prompt is used to decide whether a new term should be merged with an existing concept in the graph.

\noindent\fbox{%
  \parbox{0.98\linewidth}{%
    \small
    \textbf{System Prompt:} \\
    You are an ontology curator for a medical knowledge graph.

    Task:
    Given one NEW term and a list of existing candidate concepts, decide if the new term:
    - is the SAME concept (same meaning) as exactly one candidate, or
    - should be a NEW concept.

    Rules:
    - Treat synonyms, abbreviations, spelling variants, plural/singular, and minor wording changes as the same concept.
    - If the new term is clearly broader or narrower than all candidates in a clinically important way (e.g. "lung cancer" vs "small cell lung cancer"), treat as NEW.
    - If multiple candidates could match, choose the single best one; only use NEW when all are clearly wrong.

    Output:
    - ONE JSON object: \{ "decision": "<candidate\_id>" \} or \{ "decision": "NEW" \}.
    - No other keys, no explanations.
  }%
}

\subsection{Test Metadata Enrichment}
\label{app:prompt_test_meta}
This prompt generates cost and invasiveness metadata for new test nodes.

\noindent\fbox{%
  \parbox{0.98\linewidth}{%
    \small
    \textbf{System Prompt:} \\
    You are enriching metadata for diagnostic tests.

    Input: ONE test name.
    Output: ONE JSON object with:
    - "cost": short qualitative label of typical cost, e.g. "very low", "low", "moderate", "high", or "very high". No numeric prices.
    - "invasiveness": one of ["low", "moderate", "high"]:
    * "low"      – non-invasive, minimal risk (e.g. X-ray, blood test).
    * "moderate" – minimally invasive (needle, catheter, contrast...).
    * "high"     – clearly invasive or interventional/surgical.
    - "applicable\_population": brief description of typical use population, e.g. "adults with suspected acute coronary syndrome".

    Be approximate but clinically reasonable.
    Return ONLY the JSON with these three keys.
  }%
}

\subsection{Feature-Test Effectiveness Inference}
\label{app:prompt_ft}
This prompt determines the effectiveness of a test in confirming a specific feature.

\noindent\fbox{%
  \parbox{0.98\linewidth}{%
    \small
    \textbf{System Prompt:} \\
    You are linking features to diagnostic tests.

    Task:
    Given ONE feature and ONE test, decide how effective the test is for confirming or excluding that FEATURE itself (not any disease).

    Output: JSON with one key: "effectiveness", value in ["++", "+"].

    Meaning:
    - "++": The test directly measures or visualizes this feature with high reliability. It is the main or standard way to determine whether the feature is present.
    - "+":  The test only indirectly supports or weakly relates to this feature. It helps but is not decisive alone.

    Guidelines:
    - Judge the relation TEST $\leftrightarrow$ FEATURE, not TEST $\leftrightarrow$ disease.
    - When unsure or the link is not direct, choose "+" rather than "++".
    - Do NOT use values other than "++" or "+".

    Return ONLY: \{ "effectiveness": "++" \} or \{ "effectiveness": "+" \}.
  }%
}

\section{Prompts for Graph-Augmented Agent}
\label{app:prompts_agent}

\subsection{Information Extraction}
\label{app:prompt_ie}
This prompt extracts structured observations from the patient-doctor dialogue.

\noindent\fbox{%
\parbox{0.98\linewidth}{%
\small
\textbf{System Prompt:} \\
You are a medical information extraction assistant with expertise in clinical terminology.

Task:
Given the latest turn of conversation with a patient (which may include answers to questions, or test results), extract ONLY the NEW observations mentioned in THIS TURN.

CRITICAL: You MUST translate all observations into STANDARD MEDICAL TERMINOLOGY. Do NOT use colloquial or layperson descriptions. The extracted terms will be matched against a medical knowledge graph that uses professional clinical vocabulary.

Output a JSON object with:
\{
"observations": [
\{
"name": "...",                 // MUST use standard medical terms
"category": "symptom" | "test\_result" | "risk\_factor" | "other",
"status": "present" | "absent" | "unknown",
"source": "patient" | "test" | "other"
\}, ...
],
"tests\_mentioned": [ ... ],
"candidate\_diseases": [ ... ]
\}

Guidelines:
- ONLY include observations that are NEW or UPDATED in THIS TURN, not previous turns.
- ALWAYS convert colloquial descriptions to standard medical terminology.
- If the patient clearly denies a symptom (e.g. "No fever"), set status="absent".
- If a symptom is clearly present or strongly implied, set status="present".
- If it's unclear whether the feature is present, use status="unknown".
- Map lab/imaging findings to category="test\_result".
}%
}

\subsection{Agent Decision Making}
\label{app:prompt_decision}
This system prompt guides the final decision-making of the agent, incorporating the graph summary.

\noindent\fbox{%
  \parbox{0.98\linewidth}{%
    \small
    \textbf{System Prompt:} \\
    You are an expert diagnostician AI with access to an internal diagnostic knowledge graph.
    The graph provides:
    - relationships between diseases and features (symptoms, test results, risk factors) with typicality ("++", "+", "-", "--"),
    - relationships between features and tests, with a qualitative measure of how directly the test measures the feature ("++" or "+"),
    - simple metadata about test cost and invasiveness.

    You are interacting with a simulated patient to diagnose their condition.
    Each turn you receive NEW information from the environment (patient answers or test results).
    The system may also provide you with a GRAPH SUMMARY containing:
    - a list of structured observations parsed from the conversation,
    - a graph-based ranking of candidate diseases,
    - graph-based suggestions of useful follow-up questions and tests.

    IMPORTANT:
    - The knowledge graph may be incomplete or imperfect. Its suggestions are HINTS, not truth.
    - You must combine the graph suggestions with your own clinical reasoning and the full history.
    - Do NOT blindly follow the graph if it conflicts with clinical reasoning.
    - You have a maximum of \{max\_turns\} turns; try to be efficient and accurate.

    Available actions:
    1.  "ask"      – Ask the patient a follow-up question to gather more information.
    2.  "test"     – Order a medical test.
    3.  "diagnose" – Provide a final diagnosis if you think you have enough information.

    You MUST respond in JSON with two keys: "action\_type" and "action\_content".
  }%
}

\section{Algorithms}
\label{app:algorithms}

We provide the detailed pseudocode for the two core algorithms of \sysname{}.

\begin{algorithm}[!ht]
  \small
  \caption{Automated MDKG Construction Pipeline}
  \label{alg:construction}
  \SetAlgoLined
  \KwIn{Disease List $\mathcal{D}_{list}$, LLM $\mathcal{M}$}
  \KwOut{MDKG $\mathcal{G}$}
  Initialize empty graph $\mathcal{G}$\;
  \tcp{Stage 1: Concurrent Disease Knowledge Extraction}
  \ForEach{$d \in \mathcal{D}_{list}$ (Concurrent)}{
    $(F_d, \mathcal{T}_{map}) \leftarrow \mathcal{M}.\text{extract\_knowledge}(d)$ \tcp{Extract features \& test map}
    $P_d \leftarrow \mathcal{M}.\text{estimate\_typicality}(F_d, d)$\;
  }
  \tcp{Stage 2: Sequential Entity Alignment \& Graph Construction}
  \ForEach{$d \in \mathcal{D}_{list}$}{
    $v_d \leftarrow \text{HybridAlign}(d, \mathcal{G}.\mathcal{V})$\;
    \If{$v_d \in \mathcal{G}.\mathcal{V}_f$}{
      $\text{UpgradeNode}(v_d)$ \tcp{Promote Feature to Disease}
    }
    \ForEach{$f \in F_d$}{
      $v_f \leftarrow \text{HybridAlign}(f, \mathcal{G}.\mathcal{V}_f)$ \tcp{Hybrid Alignment}
      $\mathcal{G}.\text{add\_edge}(v_d, v_f, \text{weight}=P_d[f])$\;
      \ForEach{$\tau \in \mathcal{T}_{map}[f]$}{
        $v_\tau \leftarrow \text{HybridAlign}(\tau, \mathcal{G}.\mathcal{V}_t)$ \tcp{Test Alignment}
        $\mathcal{G}.\text{add\_edge}(v_f, v_\tau)$\;
      }
    }
  }
  \tcp{Stage 3: Metadata Enrichment \& Edge Inference}
  \ForEach{$\tau \in \mathcal{G}.\mathcal{V}_t$ (Concurrent)}{
    $\text{attr}_\tau \leftarrow \mathcal{M}.\text{enrich\_metadata}(\tau)$ \tcp{Cost/Invasiveness}
    $\mathcal{G}.\text{update\_node}(\tau, \text{attr}_\tau)$\;
  }
\end{algorithm}

\begin{algorithm}[!ht]
  \small
  \caption{Knowledge-Enhanced Diagnostic Loop}
  \label{alg:inference}
  \SetAlgoLined
  \KwIn{Dialogue History $H_t$, MDKG $\mathcal{G}$}
  \KwOut{Action $a_t$}
  \tcp{1. Info Extraction \& State Tracking}
  $O_t \leftarrow \text{ExtractObservations}(H_t)$\;
  $\text{UpdateState}(\mathcal{G}, O_t)$ \tcp{Update Confirmed/Excluded Features}
  \tcp{2. Evidence Scoring (Heuristic)}
  \ForEach{$d \in \mathcal{G}.\mathcal{V}_d$}{
    $S(d) \leftarrow \sum_{f \in O_t^+} w(d,f) - \lambda \cdot \sum_{f \in O_t^-} w(d,f)$\;
  }
  $D_{top} \leftarrow \text{TopK}(S)$\;
  \tcp{3. Utility-Based Action Suggestion}
  $A_{suggest} \leftarrow \emptyset$\;
  $T_{cand} \leftarrow \{\tau \mid \exists f \in \text{Neighbors}(D_{top}), (f, \tau) \in E_{test} \} \setminus T_{history}$\;
  \ForEach{$\tau \in T_{cand}$}{
    $V_{total} \leftarrow \sum_{f \in \text{Verifies}(\tau)} V(f) \cdot \eta(f,\tau)$\;
    $\text{Utility}(\tau) \leftarrow V_{total} / (1 + \text{Cost}(\tau) \cdot \text{Inv}(\tau))$\;
    $A_{suggest}.\text{add}(\tau, \text{Utility}(\tau))$\;
  }
  \tcp{4. LLM Decision Making}
  $S_{summary} \leftarrow \text{FormatSummary}(O_t, D_{top}, A_{suggest})$\;
  $a_t \leftarrow \text{LLM}.\text{decide}(H_t, S_{summary})$\;
  \Return $a_t$\;
\end{algorithm}

\section{Detailed Experimental Setup}
\label{app:detailed_setup}

\subsection{Environment Setup}
We developed \texttt{ClinicSim}, a modular simulation environment designed to evaluate sequential diagnostic capabilities. It integrates the dynamic interaction mechanism of \texttt{AgentClinic}~\cite{schmidgall2025agentclinicmultimodalagentbenchmark} with the cost-sensitive evaluation framework of \texttt{SDBench}~\cite{nori2025sequentialdiagnosislanguagemodels}. Specifically, \texttt{ClinicSim} consists of two core components: (1) \textbf{Patient Actor}: An LLM-driven patient agent that conducts realistic multi-turn natural language dialogues based on a complete clinical profile. It is instructed to be "passive," revealing information only when explicitly asked, to mimic the information asymmetry in real consultations; (2) \textbf{Test Result Generator}: An independent generation module responsible for returning corresponding medical test results (e.g., blood metrics, imaging descriptions) based on doctor orders. To ensure consistency, it retrieves results directly from the ground truth profile; if a specific value is missing, it returns a "normal" or "unknown" status consistent with the patient's overall condition. The maximum number of dialogue turns is set to 20. This design allows us to track the economic cost of each step in real-time, enabling a comprehensive evaluation of agent performance in the "Accuracy-Cost-Efficiency" space. The source code for \sysname{} and the \texttt{ClinicSim} environment is available at \url{https://anonymous.4open.science/r/GraphDx-BB8B}.

\subsection{Datasets}
We utilized two datasets for evaluation: (1) \textbf{MedQA Dataset:} Derived from the \texttt{medqa\_extended} dataset~\cite{schmidgall2025agentclinicmultimodalagentbenchmark}, covering diverse chief complaints, medical histories, and personality traits, totaling 200 test cases. (2) \textbf{MIMIC-IV Dataset:} To verify robustness on real-world data, we constructed 200 scenarios from the MIMIC-IV clinical database~\cite{johnsonMIMICIVFreelyAccessible2023} following the AgentClinic~\cite{schmidgall2025agentclinicmultimodalagentbenchmark} methodology. These cases contain real clinical noise, incomplete information, and complex medical histories, posing a greater challenge to the agent's information extraction and reasoning capabilities.

\subsection{Parameter Settings}
In our experiments, we set the typicality weights as $w_{++}=4.0$, $w_{+}=1.5$, $w_{-}=-1.5$, and $w_{--}=-4.0$. The penalty coefficient $\lambda$ was set to 0.6. For the utility calculation, the verification effectiveness weights were set to $w_{eff} \in \{1.0, 0.5\}$. The cost levels $C(t)$ were mapped to integers $\{1, 2, 3, 4, 5\}$ representing "very low" to "very high" costs, and the invasiveness multipliers $M_{inv}(t)$ were set to $\{1.0, 1.5, 2.0\}$ for low, moderate, and high invasiveness, respectively. The semantic retrieval threshold $\tau$ was set to 0.8.

\subsection{Baseline Settings}
To verify the generality of our method, we evaluated it on three advanced Large Language Models: (1) \textbf{DeepSeek-V3}; (2) \textbf{Kimi-k2}; (3) \textbf{Llama-3.3}. For each model, we compared three settings:
\begin{itemize}[leftmargin=*]
  \item \textbf{Standard LLM (Baseline)}: Standard agent using direct prompting for inquiry and diagnosis.
  \item \textbf{MAI-DxO~\cite{nori2025sequentialdiagnosislanguagemodels}}: The SOTA architecture proposed in SDBench, employing multi-agent collaboration (Dr. Test-Chooser, Dr. Challenger, etc.) for diagnostic orchestration.
  \item \textbf{\sysname{} (Ours)}: The proposed graph-augmented knowledge-enhanced agent.
\end{itemize}
We note that several related sequential diagnosis frameworks (e.g., DDO~\cite{jia2025ddodualdecisionoptimizationllmbased}, ACTMED~\cite{estévez2025timelyclinicaldiagnosisactive}, and DiagGym~\cite{qiu2025evolvingdiagnosticagentsvirtual}) are highly relevant in motivation, but are not included as baselines because they rely on closed-set action spaces (predefined symptom/test lists) incompatible with our open-ended natural language dialogue setting. To ensure a fair comparison, we also enforce a controlled evaluation protocol where the \texttt{ClinicSim} environment (Patient Actor, Test Result Generator, and Judge) remains fixed across all methods, isolating the impact of the agent design.

\subsection{Evaluation Metrics}
We employ three core metrics to evaluate diagnostic performance:
(1) \textbf{Diagnostic Score (0-10)}: Scored by Llama-3.3 as a judge model based on the clinical equivalence between the agent's final diagnosis and the ground truth (10: Optimal; 5: Incomplete but safe; 0: Incorrect and misleading).
(2) \textbf{Cost (\$)}: The sum of estimated costs for all medical tests ordered during the diagnosis. To ensure fair comparison, costs are estimated in USD by an independent LLM evaluator using a standardized price book prompt, rather than using the agent's internal optimization scores.
(3) \textbf{Turns}: The number of dialogue turns between the doctor and patient, reflecting diagnostic efficiency.
Note that for all metrics, we report the \textbf{Mean $\pm$ Standard Deviation} across \textbf{3 independent runs} for each scenario. The standard deviation measures the stability of the agent's performance, with randomness stemming from the LLM's decoding temperature.

\section{Hyperparameters and Constants}
\label{app:hyperparams}

We list the specific numerical values used for the qualitative attributes in our system.

\begin{table}[h]
  \centering
  \small
  \caption{Typicality Weights used in Evidence Scoring}
  \label{tab:typicality_weights}
  \begin{tabular}{lcc}
    \toprule
    Typicality Label  & Symbol & Weight Value \\
    \midrule
    Strongly Positive & ++     & 4.0          \\
    Positive          & +      & 1.5          \\
    Negative          & -      & -1.5         \\
    Strongly Negative & --     & -4.0         \\
    \bottomrule
  \end{tabular}
\end{table}

\begin{table}[h]
  \centering
  \small
  \caption{Cost Scores and Invasiveness Multipliers}
  \label{tab:cost_params}
  \begin{tabular}{lc|lc}
    \toprule
    Cost Label & Score & Invasiveness & Multiplier \\
    \midrule
    Very Low   & 1.0   & Low          & 1.0        \\
    Low        & 2.0   & Moderate     & 1.5        \\
    Moderate   & 3.0   & High         & 2.0        \\
    High       & 4.0   &              &            \\
    Very High  & 5.0   &              &            \\
    \bottomrule
  \end{tabular}
\end{table}

\section{Baseline Agent Prompts}
\label{app:prompts_baseline}

To ensure fair comparison and reproducibility, we provide the system prompts used for the Baseline Agent. The Baseline Agent uses a standard Chain-of-Thought (CoT) approach without access to the external knowledge graph.

\noindent\fbox{%
  \parbox{0.98\linewidth}{%
    \small
    \textbf{System Prompt:} \\
    You are an expert diagnostician AI. You are interacting with a patient to diagnose their condition.
    You will receive information in turns. Each turn, you will get new information, which could be the patient's response to your questions or the results of a medical test you ordered.
    Based on the information you have, you must decide on the next action to take.
    You have three possible actions:
    1.  **ask**: Ask the patient a follow-up question to gather more information.
    2.  **test**: Order a medical test to get specific data.
    3.  **diagnose**: If you have enough information, provide a final diagnosis.

    You must respond in JSON format with two keys:
    -   `"action\_type"`: One of `"ask"`, `"test"`, or `"diagnose"`.
    -   `"action\_content"`: The specific question, test name, or diagnosis.

    For example:
    If you want to ask a question:
    `\{"action\_type": "ask", "action\_content": "Have you experienced any fever?"\}`

    If you want to order a test:
    `\{"action\_type": "test", "action\_content": "Complete Blood Count"\}`

    If you are ready to diagnose:
    `\{"action\_type": "diagnose", "action\_content": "Common Cold"\}`

    You have a maximum of \{max\_turns\} turns to diagnose the patient. Strive to be efficient and accurate.
  }%
}

\section{Simulation Environment Details}
\label{app:sim_details}

We use a modular simulation environment \texttt{ClinicSim} to evaluate the agents. The environment consists of a Patient Actor and a Test Result Generator, both driven by LLMs.

\subsection{Patient Actor Prompt}
The Patient Actor simulates a real patient based on a detailed clinical profile. It is instructed to reveal information only when asked, mimicking the information asymmetry in real consultations.

\noindent\fbox{%
  \parbox{0.98\linewidth}{%
    \small
    \textbf{System Prompt:} \\
    You are role-playing as the PATIENT (or the accompanying family member IF the patient are not capable of making decisions) described below. A doctor is examining you (or your dependent).

    Rules:
    - Your profile (or your dependent's) contains very detailed medical information about you (or your dependent) but NEVER directly reveal it unless asked by the doctor.
    - NEVER directly reveal the true diagnosis to the doctor.
    - Answer the doctor's questions naturally and unprofessionally, as a real patient would.
    - ONLY share symptoms/history based on your profile when asked.
    - If unsure, say "I don't know" or give a plausible answer.

    \{profile\}
  }%
}

\subsection{Test Result Generator Prompt}
The Test Result Generator provides realistic test results based on the patient's ground truth data.

\noindent\fbox{%
  \parbox{0.98\linewidth}{%
    \small
    \textbf{System Prompt:} \\
    You are a medical test result generator. Given a test name, provide a realistic result for that test based on the patient's profile.

    Rules:
    - Return realistic values CONSISTENT with the patient's condition.
    - Return EXACTLY the required test result, NOT all results from the profile.
    - If the test is not in the profile, generate the most CONSISTENT result based on the patient's condition.
    - Format: brief, factual lab report style.
    - NO diagnosis, NO interpretation, NO conversation.

    \{profile\}
  }%
}

\subsection{Evaluation Prompts}
\label{app:eval_prompts}

To ensure consistent and scalable evaluation, we utilize LLM-based judges for both diagnostic accuracy scoring and cost estimation.

\subsubsection{Diagnostic Accuracy Evaluator}
We employ an LLM-based evaluator to grade the diagnostic outcome against the ground truth.

\noindent\fbox{%
  \parbox{0.98\linewidth}{%
    \small
    \textbf{System Prompt:} \\
    You are an expert medical evaluator. Compare the agent's diagnosis with the ground truth diagnosis. Provide a score from 0 to 10. Use this scale:
    \begin{itemize}[leftmargin=*]
        \item 10 (Optimal): The agent's final diagnosis is clinically equivalent to the ground truth diagnosis.
        \item 5 (Acceptable): The agent's diagnosis is incorrect/incomplete, BUT it is safe and does not mislead future treatment (e.g., a correct referral, or a safe differential).
        \item 0 (Harmful): The agent's diagnosis is incorrect AND misleading, potentially causing harm or treatment delay.
    \end{itemize}
    Above descriptions are anchor points; intermediate scores (1-4, 6-9) should reflect gradations in clinical relevance and safety.

    You MUST output a JSON object with ONLY the "score" field. No other fields or text allowed.

    Example output:
    \{``score'': 8\}

    Full profile of the case:
    \{profile\_text\}

    Agent's Diagnosis:
    \{agent\_diagnosis\}
  }%
}

\subsubsection{Cost Estimation Methodology \& Prompt}
\label{sec:cost_estimation}
To evaluate the economic efficiency of diagnostic agents, we calculate the financial cost of each diagnostic session. Since real-world pricing varies continuously, we employ a Large Language Model (specifically \texttt{Llama-3.3-70b-instruct}) acting as a "Medical Billing Expert" to estimate the cost of each ordered test. The model is prompted to provide realistic US-based pricing (in USD) for each specific test name.

\noindent\fbox{%
  \parbox{0.98\linewidth}{%
    \small
    \textbf{System Prompt:} \\
    You are a medical billing expert. For each of the following medical tests, provide an estimated cost in USD. Use realistic pricing based on typical costs in the United States healthcare system.

    \textbf{CRITICAL RULES:}
    \begin{enumerate}[leftmargin=*]
        \item You MUST present the output as a valid JSON object where keys are the test names and values are the estimated costs (as numbers).
        \item Do not include any other text or explanations.
        \item ONLY include tests that are explicitly listed below. Do NOT add any tests that are not in the list.
    \end{enumerate}

    Example output when tests are ordered:
    \{
      ``Complete Blood Count (CBC)'': 50,
      ``Basic Metabolic Panel (BMP)'': 75
    \}

    Tests Ordered:
    \{ordered\_tests\}
  }%
}

\section{Additional Experimental Analysis}
\label{app:additional_analysis}

\subsection{MDKG Statistics}
\label{app:graph_stats}
Before evaluating the diagnostic performance, we assessed the quality of the constructed MDKG through a rigorous manual verification process. We randomly sampled 243 triples (disease-feature associations) from the generated graph and asked medical professionals to classify them into three categories: \textit{Correct} (medically accurate and precise), \textit{Plausible} (reasonable but potentially vague or context-dependent), and \textit{Incorrect} (medically false).

The evaluation results yielded 186 (76.5\%) \textit{Correct} triples, 37 (15.2\%) \textit{Plausible} triples, and 20 (8.2\%) \textit{Incorrect} triples. The overall validity rate (Correct + Plausible) reached 91.8\%, demonstrating that our automated pipeline can reliably extract high-quality medical knowledge. The small fraction of incorrect triples mostly involved over-generalized associations (e.g., linking generic symptoms to specific rare diseases), which are often filtered out during the subsequent evidence scoring phase.

\begin{figure*}[htbp]
  \centering
  \begin{subfigure}{0.32\linewidth}
    \includegraphics[width=\linewidth]{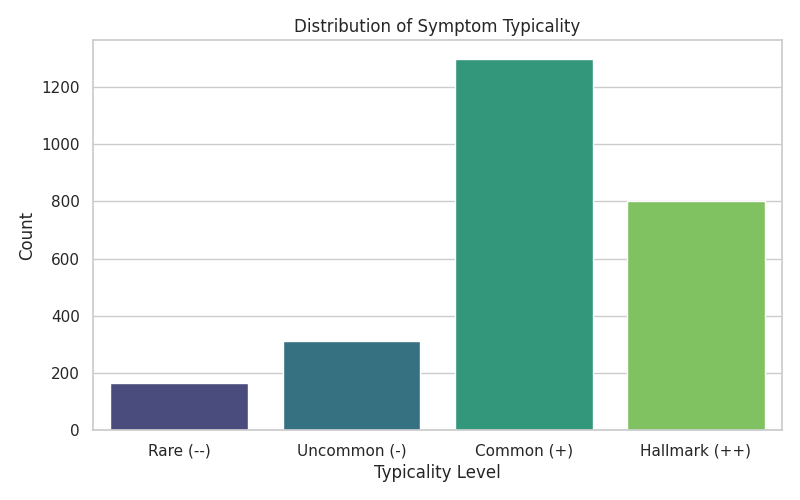}
    \caption{Feature Typicality Distribution}
    \label{fig:graph_stats_typ}
  \end{subfigure}
  \hfill
  \begin{subfigure}{0.32\linewidth}
    \includegraphics[width=\linewidth]{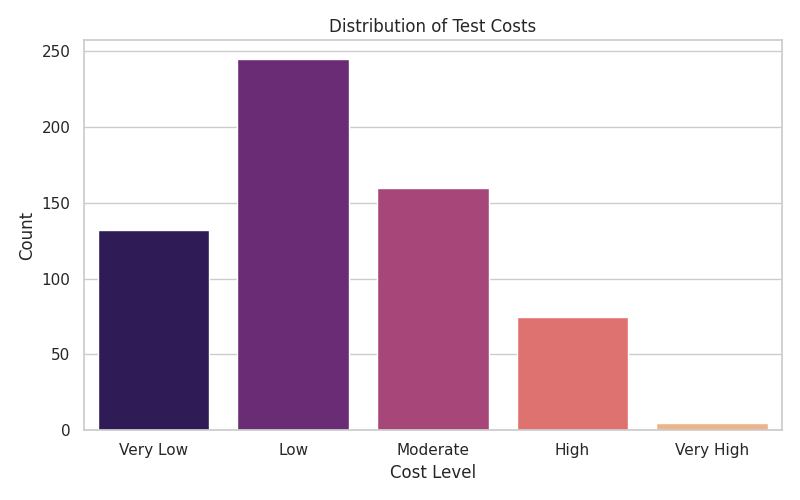}
    \caption{Test Cost Distribution}
    \label{fig:graph_stats_cost}
  \end{subfigure}
  \hfill
  \begin{subfigure}{0.32\linewidth}
    \includegraphics[width=\linewidth]{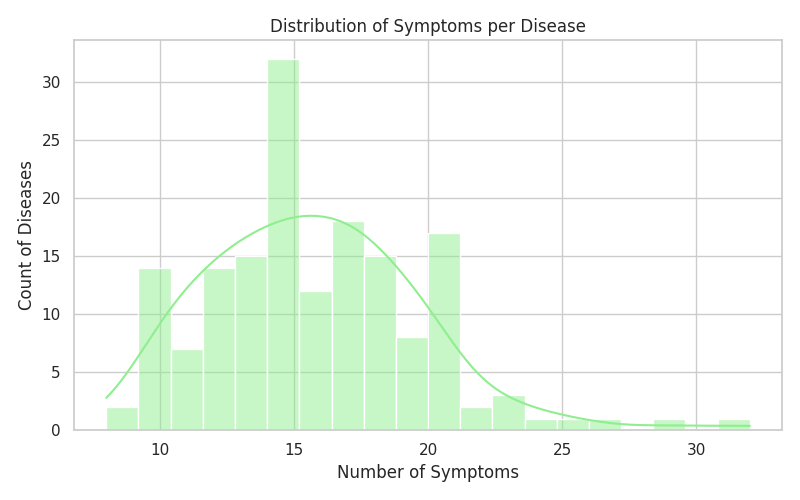}
    \caption{Disease-Feature Degree Dist.}
    \label{fig:graph_stats_degree}
  \end{subfigure}
  \caption{Statistical distribution of key attributes in MDKG.}
  \label{fig:graph_stats}
\end{figure*}

Figure~\ref{fig:graph_stats} shows the distribution of key graph attributes. Figure~\ref{fig:graph_stats_typ} displays the distribution of feature typicality ($P(f|d)$), presenting a reasonable skewed distribution, indicating that the graph can distinguish between "hallmark features" (high typicality) and "common features." Figure~\ref{fig:graph_stats_cost} shows the distribution of test costs, covering a wide range from low-cost (e.g., CBC) to high-cost (e.g., MRI) tests, providing a basis for cost-sensitive planning. Figure~\ref{fig:graph_stats_degree} illustrates the degree distribution of disease nodes (i.e., number of features per disease), with an average of 15.7 features per disease, providing sufficient evidence for differential diagnosis.

\subsection{Parameter Sensitivity Analysis}
\label{app:sensitivity}
To assess the robustness of our evidence scoring mechanism while avoiding the prohibitive computational cost of re-running full agent simulations, we conducted an offline sensitivity analysis on the Llama-3.3 interaction logs. We re-evaluated the disease ranking logic by varying two key hyperparameters: the weight of strong positive evidence ($w_{++}$) and the penalty coefficient for missing expected features ($\lambda$).

\begin{figure}[htbp]
  \centering
  \includegraphics[width=\linewidth]{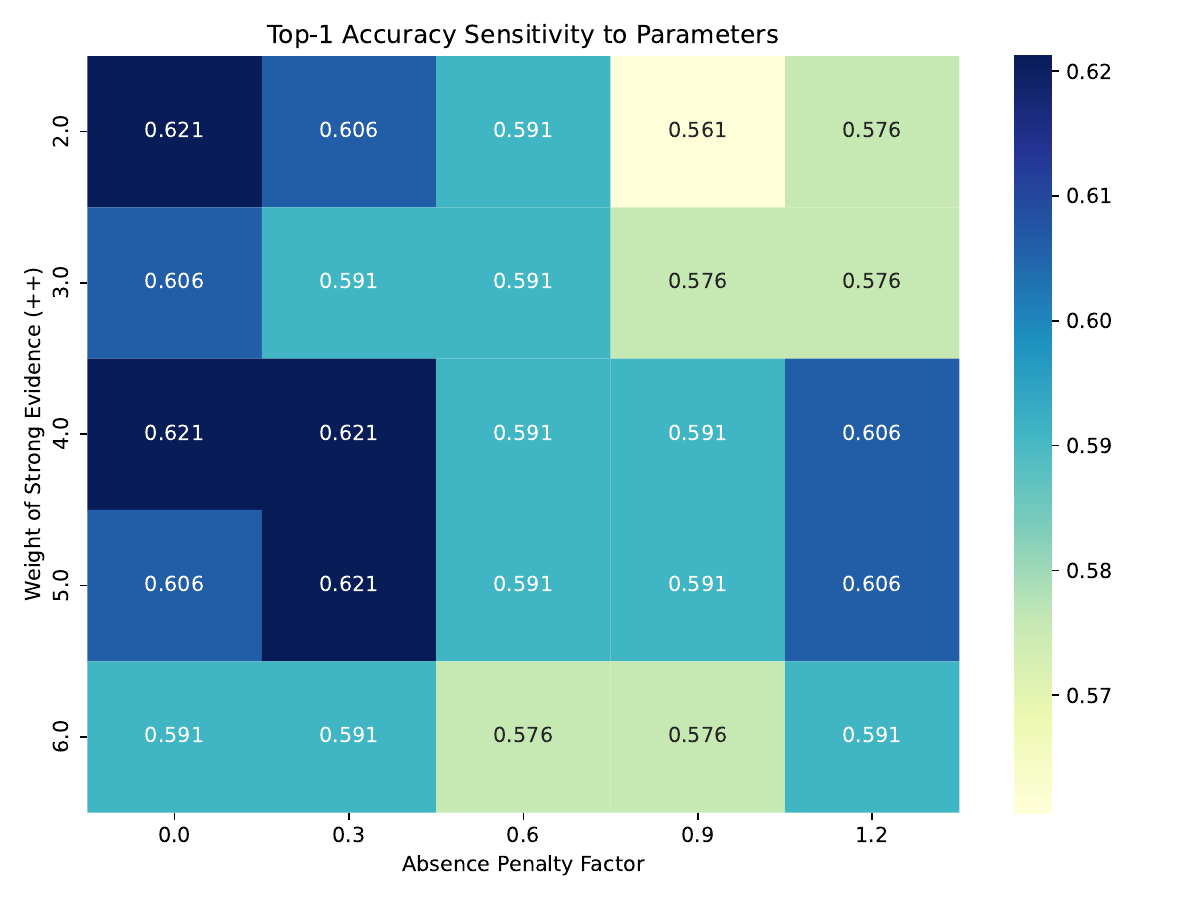}
  \caption{Sensitivity of Top-1 Accuracy to variations in Strong Evidence Weight ($w_{++}$) and Absence Penalty ($\lambda$). The system exhibits a stable performance plateau, demonstrating robustness.}
  \label{fig:sensitivity}
\end{figure}

Figure~\ref{fig:sensitivity} illustrates the Top-1 accuracy of the ground truth disease across a grid of parameter settings. We observe that:
\begin{itemize}[leftmargin=*]
  \item \textbf{Robustness to Weight Variations:} The system maintains stable performance (Top-1 accuracy $\approx 60\%$, Top-5 accuracy $\approx 88\%$) across a wide range of $w_{++}$ values (2.0--5.0). This indicates that the ranking is primarily driven by the structural correctness of the graph (presence of correct edges) rather than fine-tuned parameters.
  \item \textbf{Impact of Absence Penalty:} A lower absence penalty ($\lambda \in [0.0, 0.3]$) yields slightly better results than higher penalties. This suggests that in natural language dialogue, "missing" features are often simply unmentioned by the patient rather than clinically absent, so the system should be cautious about penalizing them too heavily.
\end{itemize}

\subsection{Case Study}
\label{app:case_study}
To deeply understand the decision-making advantages of \sysname{} in different clinical contexts, we selected three representative cases from the MedQA-Extended dataset for detailed comparative analysis. These cases reveal typical failure modes of baseline models in \textbf{rare disease reasoning}, \textbf{anatomical logic}, and \textbf{cost-effective diagnosis}, while \sysname{} successfully avoids these pitfalls through structured knowledge.

\begin{figure}[htbp]
  \centering
  \includegraphics[width=\linewidth]{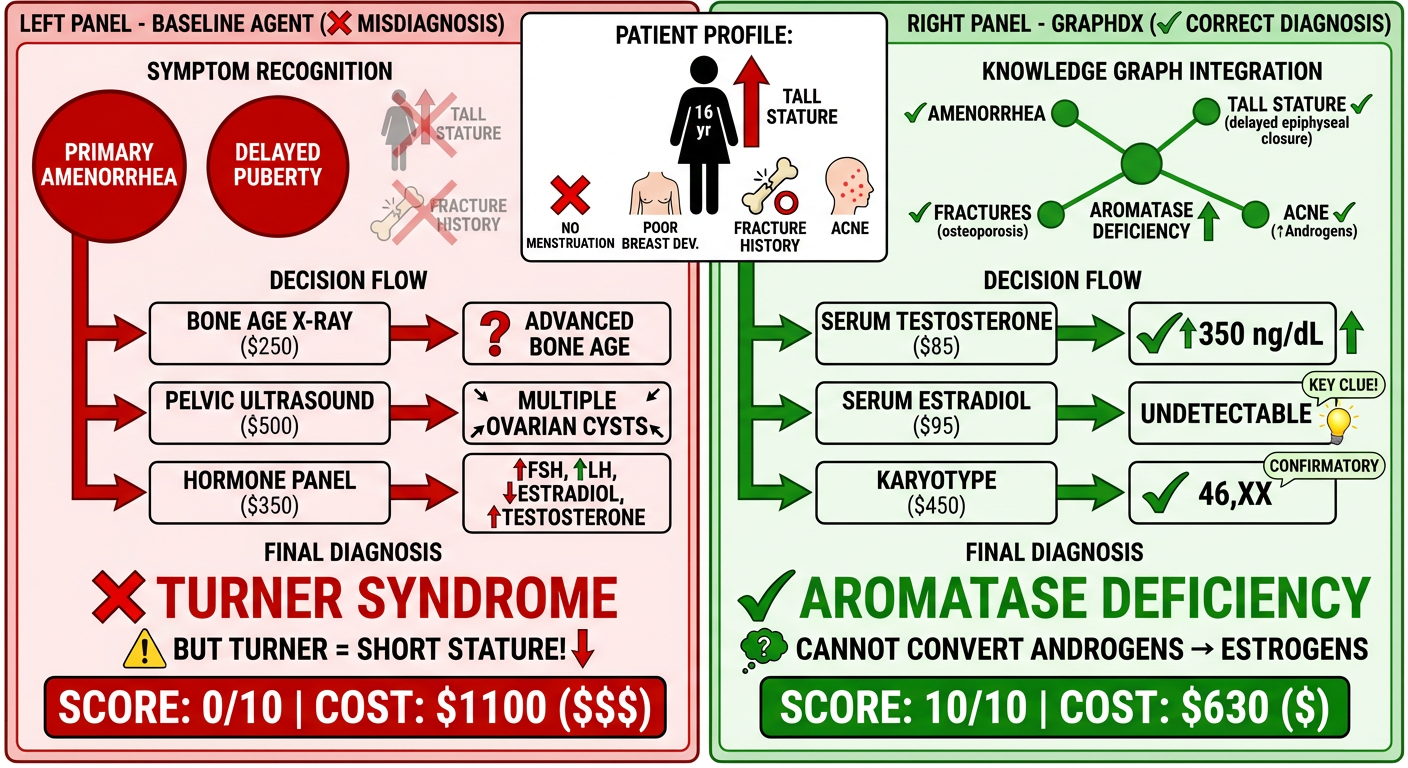}
  \caption{Typical Case Analysis (Scenario 50: Primary Amenorrhea). The Baseline agent (left) misdiagnosed \textbf{Turner Syndrome} by focusing on amenorrhea while ignoring the contradictory evidence of \textbf{tall stature} and \textbf{fractures}. In contrast, \sysname{} (right) successfully linked these cues to \textbf{Aromatase Deficiency} via the structured graph, demonstrating the navigational role of structured knowledge.}
  \label{fig:case_study}
\end{figure}

\paragraph{Case 1: Capturing Subtle Clues in Rare Diseases (Scenario 50)}
As shown in Figure~\ref{fig:case_study}, \textbf{Case Background:} A 16-year-old female presented with primary amenorrhea and delayed breast development. Notably, her history included "fractures from minor falls" and she was "tall for her age".
\begin{itemize}[leftmargin=*]
  \item \textbf{Baseline:} Focused solely on amenorrhea and delayed puberty, diagnosing "Turner Syndrome". It failed to recognize that Turner Syndrome typically presents with \textit{short} stature, contradicting the patient's "tall" description. It also ignored the fracture history.
  \item \textbf{\sysname{}:} The graph reasoning engine successfully linked "fractures" (osteoporosis) and "tall stature" (delayed epiphyseal closure) with "amenorrhea". It identified the rare etiology \textbf{"Aromatase Deficiency"} (inability to synthesize estrogen from androgens) and confirmed it via hormone panel and karyotype, achieving a precise diagnosis.
\end{itemize}

\paragraph{Case 2: Strict Anatomical and Functional Constraints (Scenario 14)}
\textbf{Case Background:} A patient reported a swollen right ring finger after a football injury, stating, "I can't bend the tip of my finger at all when I try to make a fist."
\begin{itemize}[leftmargin=*]
  \item \textbf{Baseline:} Confused the functional description, diagnosing "Mallet Finger" (an extensor tendon injury characterized by inability to \textit{extend} the tip).
  \item \textbf{\sysname{}:} The MDKG explicitly encodes the functional logic of musculoskeletal anatomy. It correctly interpreted the inability to \textit{flex} the distal joint as a rupture of the Flexor Digitorum Profundus tendon, correctly diagnosing \textbf{"Jersey Finger"}.
\end{itemize}

\paragraph{Case 3: Cost-Effective Decision Making (Scenario 66)}
\textbf{Case Background:} A 1-month-old preterm infant presented with "apnea episodes" during sleep and "pale skin".
\begin{itemize}[leftmargin=*]
  \item \textbf{Baseline:} Reacted defensively to the high-risk symptom "apnea" by ordering an expensive \textbf{Polysomnography (\$2500)}, diagnosing "Apnea of Prematurity" without investigating the root cause.
  \item \textbf{\sysname{}:} Integrated the "pale skin" clue with "apnea" in the evidence-weighted graph, hypothesizing anemia as a potential cause. It prioritized a low-cost \textbf{Complete Blood Count (\$50)}, revealing low hemoglobin (8.2 g/dL), and correctly diagnosed \textbf{"Anemia of Prematurity"}. This demonstrates how utility-based planning avoids unnecessary expensive testing.
\end{itemize}

\end{document}